\documentclass{article}

\usepackage[nonatbib, preprint]{neurips_2020}

\usepackage[utf8]{inputenc} 
\usepackage[T1]{fontenc}    
\usepackage{hyperref}       
\usepackage{url}            
\usepackage{booktabs}       
\usepackage{amsfonts}       
\usepackage{nicefrac}       
\usepackage{microtype}      

\usepackage{helvet}  
\usepackage{courier}  
\usepackage{adjustbox}
\usepackage{xcolor}
\usepackage{amsmath,amssymb}

\usepackage{graphicx} 
\usepackage[hang]{subfigure} 
\graphicspath{{figures/}}
\usepackage{wrapfig}
\usepackage{algorithm}
\usepackage{algorithmic}

\usepackage{multirow}

\usepackage{breakcites}

\allowdisplaybreaks[1]

\ifx\proof\undefined
\newenvironment{proof}{\par\noindent{\bf Proof\ }}{\hfill\BlackBox\\[2mm]}
\fi

\ifx\theorem\undefined
\newtheorem{theorem}{Theorem}
\fi
\ifx\example\undefined
\newtheorem{example}{Example}
\fi
\ifx\lemma\undefined
\newtheorem{lemma}[theorem]{Lemma}
\fi

\ifx\corollary\undefined
\newtheorem{corollary}[theorem]{Corollary}
\fi

\ifx\assumption\undefined
\newtheorem{assumption}{Assumption}
\fi

\ifx\definition\undefined
\newtheorem{definition}{Definition}
\fi

\ifx\proposition\undefined
\newtheorem{proposition}[theorem]{Proposition}
\fi

\ifx\remark\undefined
\newtheorem{remark}{Remark}
\fi

\ifx\conjecture\undefined
\newtheorem{conjecture}[theorem]{Conjecture}
\fi

\ifx\factoid\undefined
\newtheorem{factoid}[theorem]{Fact}
\fi

\ifx\axiom\undefined
\newtheorem{axiom}[theorem]{Axiom}
\fi

\newcommand{\RN}[1]{%
	\textup{\lowercase\expandafter{\it \romannumeral#1}}%
}

\newcommand{\ROM}[1]{\uppercase\expandafter{\romannumeral #1\relax}}

\input{Definitions}

\title{KernelNet: A Data-Dependent Kernel Parameterization for Deep Generative Modeling}

\author{
		Yufan Zhou,\quad  Changyou Chen,\quad Jinhui Xu \\
		State University of New York at Buffalo\\
		\texttt{\{yufanzho, changyou, jinhui\}@buffalo.edu} \\
}

\begin{document} 
\maketitle
\begin{abstract} 
    Learning with kernels is an important concept in machine learning. Standard approaches for kernel methods often use predefined kernels that require careful selection of hyperparameters. To mitigate this burden, we propose in this paper a framework  to construct and learn a data-dependent kernel based on random features and implicit spectral distributions that are parameterized by deep neural networks. The constructed network (called {\em KernelNet}) can be applied to deep generative modeling in various scenarios, including  two popular learning paradigms in deep generative models, MMD-GAN and implicit Variational Autoencoder (VAE).  We show that our proposed kernel indeed exists in applications and is guaranteed to be positive definite. 
    Furthermore, the induced Maximum Mean Discrepancy (MMD) can endow the continuity property in weak topology by simple regularization. Extensive experiments indicate that our proposed KernelNet consistently achieves better performance compared to related methods. 
\end{abstract}

\section{Introduction}

Kernels are important tools in machine learning, and can be used in a wide range of applications. For example, support vector machine (SVM) \cite{DBLP:books/lib/ScholkopfS02} can perform efficient non-linear classification task based on non-linear mappings through kernels; MMD-GAN  \cite{DBLP:conf/nips/LiCCYP17} can handle image generation task by utilizing Maximum Mean Discrepancy (MMD) \cite{DBLP:journals/jmlr/GrettonBRSS12}. Other kernel-based methods such as those in \cite{DBLP:conf/icml/YinZ18,FengWL:UAI17} use kernels for  estimating quantities like gradients. These models are built by either restricting the solution space to a Reproducing Kernel Hilbert Space (RKHS) induced by a kernel, or adopting the MMD as the objective functions that require specified kernels in their MMDs.

A 
not-so-desirable issue 
of the aforementioned kernel-based methods, however, is the need of selecting appropriate kernels and hyper-parameters. Such selections are critical in obtaining good performance, and manual selection often leads 
to sub-optimal solutions. Some previous works have tried to mitigate this problem. For example, \cite{DBLP:journals/jmlr/GonenA11} suggests to learn a combination of some predefined kernels; \cite{ong2004learning} proposes to relax the restriction of positive definiteness, which leads to a richer family of kernels. Alternatively, some other recent works focus on learning kernels based on random features \cite{RahimiR:NIPS07,DBLP:conf/eccv/BazavanLS12,DBLP:conf/icml/WilsonA13, li2019implicit} (see Section~\ref{sec:kernelp} for a more detailed description). 

In this paper, we propose a new kernel-learning paradigm by formulating the kernel as an expectation w.r.t.\! learnable random features. These random features are sampled from an expressive distribution of the corresponding kernel in the spectral domain (which is called spectral distribution). Specifically, we propose to parameterize the spectral distribution as a data-dependent distribution, meaning that it depends on the input data of the kernel function. The data-dependent distribution is represented by a deep neural network (DNN), which outputs samples following the distribution. We call the resulting network {\em KernelNet}, and the kernel corresponding to the data-dependent distribution a {\em data-dependent kernel}. 

The work that is most related to ours is perhaps the one in \cite{li2019implicit}.
It models the spectral distribution as a data-independent distribution, and in some sense can be seen as a special case of ours. Due to the added data-dependent component in KernelNet, our method is thus more general and more expressive. It often 
leads to performance improvement over the data-independent parameterization, as evidenced by our experiments.

Our proposed KernelNet can be readily applicable to a number of 
existing models. As an example, we show that it can be applied to  two representative Deep Generative Models (DGMs): Generative Adversarial Network (GAN) \cite{GoodfellowAMXWOCB:NIPS14} and Variational Autoencoder (VAE) \cite{KingmaW:ICLR14,pmlr-v32-rezende14}. Specifically, $\RN{1})$ we apply our proposed kernel to several variants of MMD-GAN
. We show that our proposed method leads to better performance and the induced MMD can easily satisfy the continuity in weak topology, which is an important property to for stable optimization procedures. $\RN{2})$ We propose an implicit VAE model, where an MMD-regularizer is incorporated into the objective function  of VAE. Our model is implicit in the sense that our posterior distribution is parameterized as an expressive distribution without a closed form, which is different from the typical Gaussian assumption in standard VAE and thus enables us to model a much more flexible latent space. 
To summarize, our paper has the following contributions:
\begin{itemize}
    \vspace{-0.05in}
    \item We introduce the concept of data-dependent kernel, whose spectral distribution depends on the input pair of kernel. We prove the existence and the positive definiteness of the proposed data-dependent kernel, and present a practical way to construct such a 
    kernel.
    \vspace{-0.03in}
    \item We show that our proposed kernel can be applied to two popular deep generative models: GAN and VAE, and how the MMD in our proposed GAN satisfies the continuity property in weak topology.
    \vspace{-0.03in}
    \item Extensive experiments suggest that our proposed kernel can lead to better performance compared to pre-defined kernel and previous representative kernel learning method \cite{li2019implicit}.
\end{itemize}
\section{Preliminaries}
We start by reviewing MMD-GAN and Info-VAE, two DGMs where our proposed method apply.
\subsection{MMD-GAN}
GAN is one of the most popular and powerful generative models in deep learning. It consists of a generator and a discriminator. The generator generates samples by transforming a simple noise distribution to an implicit distribution $\mathbb{Q}$, where one can easily generate samples from this distribution, but the density function is unknown. The discriminator is trained to distinguish the true training data distribution $\mathbb{P}$ and the implicit distribution $\mathbb{Q}$ induced by the generator. The generator, on the other hand, is trained to fool the discriminator. At the equilibrium, the generator should be able to generate samples that are distributed as the true data distribution $\mathbb{P}$.

MMD-GAN achieves this by miminzing the maximum mean discrepancy (MMD) between two probability measures, the data and model distributions. The MMD between two probability distributions $\mathbb{P}$ and $\mathbb{Q}$ is defined as:
\[
    \textup{MMD}_{k}(\mathbb{P}, \mathbb{Q}) = \sup _{f:\Vert f\Vert_{\mathcal{H}} \leq 1}\mathbb{E}_{\xb \sim \mathbb{P}}[f(\xb)] - \mathbb{E}_{\yb \sim \mathbb{Q}}[f(\yb)],
\]
where $\mathcal{H}$ is a Reproducing Kernel Hilbert Space (RKHS) and $f$ is a function in this RKHS. 

For an RKHS induced by kernel $k$, MMD can be computed using the following equation: 
\begin{align*}
    \textup{MMD}^2_{k}(\mathbb{P}, \mathbb{Q}) = \mathbb{E}_{\xb,\xb' \sim \mathbb{P}}[k(\xb, \xb')] - 2\mathbb{E}_{\xb \sim \mathbb{P}, \yb \sim \mathbb{Q}}[k(\xb, \yb)] + \mathbb{E}_{\yb,\yb' \sim \mathbb{Q}}[k(\yb, \yb')].
\end{align*}
For a characteristic kernel, $\textup{MMD}_{k}(\mathbb{P}, \mathbb{Q})=0$ if and only if $\mathbb{P} = \mathbb{Q}$. Thus, MMD can be used as a way of measuring the similarity of distributions or as a training objective. 

\cite{DBLP:conf/nips/LiCCYP17} propose to define the kernel as a composition of an injective function $h_{\phib}$ for feature extraction and a kernel function $k$ for kernel evaluation, {\it e.g.}, $k_{\phib} = k \circ h_{\phib}$. $k_{\phib}$ is also a valid kernel function \cite{10.5555/975545}. For example, if $k$ is the RBF kernel, $k_{\phib}(\xb, \yb) = \exp(-\Vert h_{\phib}(\xb) - h_{\phib}(\yb) \Vert^2)$ is also a valid kernel.  

Denote $f_{\thetab}$ as a generator parameterized by $\thetab$. Let $\mathbb{P}$ represent the training data distribution and $\mathbb{Q}$ the implicit distribution induced by the generator. The objective of MMD-GAN is formulated as:
\[
    \min_{\thetab} \max_{\phib} \text{MMD}^2_{k_{\phib}}(\mathbb{P}, \mathbb{Q}).
\]
Because of the min-max adversarial training, $\mathbb{Q}$ will eventually match $\mathbb{P}$ in theory. However, MMD-GAN still suffers from training instability. It has been shown that better performance can be achieved by defining variants of MMD as objective functions. 

\cite{DBLP:conf/nips/ArbelSBG18} proposes to replace the objective function of MMD-GAN by the Scaled Maximum Mean Discrepancy (SMMD), which leads to the SMMD-GAN. The SMMD is defined as:
\begin{align*}
    \textup{SMMD}_{\phib,\lambda}(\mathbb{P}, \mathbb{Q})&:= 
    %
    \sigma_{\phib,\lambda}\textup{MMD}_{k_{\phib}}(\mathbb{P}, \mathbb{Q}), 
\end{align*}
{\small
\begin{align*}
        \text{where }\sigma_{\phib,\lambda}:= \bigg\{ \lambda + \int k(\xb, \xb) d\mathbb{P}(\xb) + \sum_{i=1}^d \int \dfrac{\partial^2k(\yb_1,\yb_2)}{\partial \yb_{1i}\partial \yb_{2i}}\vert_{(\yb_1, \yb_2)=(\xb, \xb)} d\mathbb{P}(\xb)\bigg\}^{-1/2},
\end{align*}
}
and $d$ is the dimensionality of the data; $\yb_i$ denotes the $i^{th}$ element of $\yb$; $\lambda$ is a hyper-parameter. 

\cite{wang2018improving} propose a repulsive loss function for the discriminator in MMD-GAN, which is defined as:
\begin{align*}
    L_{\eta, \phib} = \eta\mathbb{E}_{\xb, \xb' \sim \mathbb{P}}\left[ k_{\phib}(\xb, \xb')\right] - \mathbb{E}_{\yb,\yb' \sim \mathbb{Q}}[k_{\phib}(\yb, \yb')]  - (\eta-1)\mathbb{E}_{\xb \sim \mathbb{P}, \yb \sim \mathbb{Q}}[k_{\phib}(\xb, \yb)].
\end{align*}
Intuitively, the repulsive loss will explore the differences among data, leading to better performance in the data generation tasks.
\subsection{Info-VAE}

VAE and its variants are another family of DGMs where latent spaces define the posterior distributions. Specifically, define a generative process for an observation $\xb \in \mathbb{R}^D$, starting from the corresponding latent variable $\zb\in\mathbb{R}^d$, as: $\xb | \zb \sim p_{\thetab}(\xb|\zb)$ with $\zb \sim p(\zb)$, where $p(\zb)$ is called the prior distribution. Transformation from $\zb$ to $\xb$ is performed using a neural network parameterized by $\thetab$, which is called the decoder. For efficient inference of $\zb$,  VAE \cite{KingmaW:ICLR14} defines an inference network (or encoder) to generate $\zb$ from $\xb$, with the corresponding distribution being $q_{\phib}(\zb | \xb)$ parameterized by $\phib$ (also called the variational distribution or variational posterior distribution). 

VAE is optimized by maximizing the Evidence Lower Bound (ELBO) $\mathbb{E}_{q_{\phib}(\zb | \xb)}\left[\text{log}p_{\thetab}(\xb \vert \zb)\right] - \textup{KL}[q_{\phib}(\zb | \xb)\Vert p(\zb)]$, which can be understood as simultaneously reconstructing the observations and minimizing the Kullback-Leibler (KL) divergence between prior and posterior distributions.

Info-VAE \cite{DBLP:journals/corr/ZhaoSE17b} is a generalization of VAE by introducing an information-theoretic regularizer into the VAE framework. The objective of Info-VAE is:
\begin{align}\label{eq:infovae_obj}
    \max_{\phib, \thetab} - \mathbb{E}_{q_{\phib}(\zb)}\{\textup{KL}[q_{\phib}(\xb \vert \zb)\Vert p_{\thetab}(\xb \vert \zb)]\} + \alpha\mathbf{I}_{q_{\phib}}(\xb, \zb)  -\lambda \textup{KL}[q_{\phib}(\zb)\Vert p(\zb)],
\end{align}
where $\mathbf{I}_{q_{\phib}}(\xb, \zb) = \mathbb{E}_{q(\xb)}\{\text{KL}[q_{\phib}(\zb \vert \xb)\Vert q_{\phib}(\zb)]\}$ is the mutual information between $\xb$ and $\zb$.

\section{KernelNet for Learning Deep Generative Models}

\subsection{The Proposed KernelNet}\label{sec:kernelp}

To describe our construction, we start with a classic result on positive definite functions \cite{Rudin:94}, 
which states that a continuous function $\kappa(\zb)$ in $R^d$ is positive definite if and only if it is the Fourier transform of a non-negative measure \cite{Rudin:94}.
Let $\zeta_{\omegab}(\zb) = e^{j\omegab^\intercal \zb}$. 
Based on  
\cite{RahimiR:NIPS07,Rudin:94}, a kernel such that $\kappa(\zb_1, \zb_2)=\tilde{\kappa}(\zb_1 - \zb_2)$ can be represented as:
\begin{align}\label{eq:kernel}
    \kappa(\zb_1, \zb_2) &= \int _{R^d}p(\omegab)e^{j\omegab^\intercal(\zb_1-\zb_2)}d\omegab = \mathbb{E}_{\omegab}[\zeta_{\omegab}(\zb_1)\zeta_{\omegab}(\zb_2)^*]~,
\end{align}
\noindent where $j$ is an indeterminate satisfying $j^2=-1$, and ``*'' denotes the conjugate transpose. 
The kernel representation \eqref{eq:kernel} directly allows us to construct an unbiased estimator for $\kappa(\cdot, \cdot)$ by introducing any valid distribution $p(\omegab)$ for the augmented variable $\omegab$, called the {\em spectral distribution}. 
In the following, we first reformulate \eqref{eq:kernel} into two equivalent forms for the purposes of analysis and algorithm design, respectively. Because the probability density function and kernel function are real-valued, by Euler's formula, we can rewrite the kernel as in Theorem~\ref{pro:kernel_cos}. Proofs for all the theoretical results in this paper will be deferred to the appendix.
\begin{theorem}\label{pro:kernel_cos}
    Let $\omegab$ be drawn from some spectral distribution $p(\omegab)$, and $b$ be drawn uniformly from $[0, 2\pi]$. The real-valued kernel in \eqref{eq:kernel} can be reformulated into the following two forms:
    \begin{flalign}
        \kappa(\zb_1, \zb_2) = \mathbb{E}_{\omegab \sim p(\omegab)}\{\cos\left[\omegab^\intercal(\zb_1-\zb_2)\right]\}
        = \mathbb{E}_{\omegab, b}\left[2\cos(\omegab^\intercal\zb_1 +b)\cos(\omegab^\intercal\zb_2+b) \label{eq:kernel_cos} \right],
    \end{flalign}
    where $\omegab \sim p(\omegab)$ and $b\sim \mathcal{U}[0, 2\pi]$.
\end{theorem}
In the above two representations, the first one is more convenient for theoretical analysis, and the second one is found more stable in algorithmic implementation. 
To enhance the expressive power, we can make the distribution $p(\omegab)$ complex enough and learnable by parameterizing it with a DNN that induces an implicit distribution. Specifically, we rewrite $p(\omegab)$ as $p_{\psib_1}(\omegab)$ with parameter $\psib_1$. A sample $\omegab$ from $p_{\psib_1}$ is modeled as the following generating process:
\begin{equation}\label{eq:kernel_part_1}
    \omegab_{\psib_1} = g_{\psib_1}(\epsilonb),
\end{equation} 
where $\epsilonb$ is sampled from simple distribution such as uniform distribution $\mathcal{U}[\mathbf{-1}, \mathbf{1}]$ or standard Gaussian $\mathcal{N}(\textbf{0}, \textbf{I})$, thus $g_{\psib_1}(\epsilon)$ denotes the output of a DNN parameterized by $\psib_1$ with the input $\epsilonb$ drawn from some simple distributions.

\paragraph{From data-independent to data-dependent kernels}
Although the above kernel parameterization is flexible to represent a rich family of implicit spectral distributions, it can be further extended by introducing a data-dependent spectral distribution. By data-dependent spectral distribution, we mean that there are some kernels satisfying \eqref{eq:kernel}, whose spectral distributions $p(\omegab)$ depend on the data pair $(\zb_1, \zb_2)$, {\it i.e.}, there exists a $p(\omegab\vert \zb_1, \zb_2)$ for each pair $(\zb_1, \zb_2)$. 
We first prove the following theorem, which serves as the foundation of our proposed method to guarantee the positive definiteness of a data-dependent kernel. 

\begin{theorem}\label{thm:data_dependent_postive_definite}
    Let $\omegab$ be drawn from a data-dependent spectral distribution $p(\omegab\vert \zb_1, \zb_2)$. If the probability density function can be formulated as 
    $p(\omegab\vert \zb_1, \zb_2) = r(\omegab)s(\omegab, \zb_1)s(\omegab, \zb_2)$,
    where $r$ and $s$ are functions such that $r(\omegab) \geq 0$  ($\forall \omegab$) and $s(\omegab, \zb_1)s(\omegab, \zb_2) \geq 0$ ($\forall (\omegab, \zb_1, \zb_2)$). Then, the kernel $\kappa(\zb_1, \zb_2) = \int _{R^d}p(\omegab \vert \zb_1, \zb_2)e^{j\omegab^\intercal(\zb_1-\zb_2)}d\omegab $ is positive definite.
\end{theorem}
Theorem \ref{thm:data_dependent_postive_definite} implies that given any spectral distribution $p(\omegab)$, we can always construct a data-dependent positive definite kernel through $p(\omegab)s(\omegab, \zb_1)s(\omegab, \zb_2)$ by using a non-constant function $s$. 
We note that $p(\omegab\vert \zb_1, \zb_2) = r(\omegab)s(\omegab, \zb_1)s(\omegab, \zb_2)$ is a simple and natural condition for 
data-dependency. It indicates that $\zb_1$ and $\zb_2$ influence the distribution $p(\omegab\vert \zb_1, \zb_2)$ through the same function $s$, which also guarantees that the resulting kernel $\kappa_{\psib_2}(\zb_1, \zb_2) = \int _{R^d}p_{\psib_2}(\omegab \vert \zb_1, \zb_2)e^{j\omegab^\intercal(\zb_1-\zb_2)}d\omegab $ is symmetric i.e. $ \kappa_{\psib_2}(\zb_1, \zb_2) = \kappa_{\psib_2}(\zb_2, \zb_1)$ because $p(\omegab\vert \zb_1, \zb_2) = p(\omegab\vert \zb_2, \zb_1)$. We will show that in our construction of the KernelNet below, this condition is satisfied.  Moreover, if $s(\omegab, \zb_1)s(\omegab, \zb_2) = c$ for some constant $c$, $\zb_1$ and $\zb_2$ will not influence the spectral distribution $p(\omegab\vert \zb_1, \zb_2)$, in which case the resulting kernel is data-independent as in \eqref{eq:kernel_part_1} \cite{li2019implicit}. 

We use the term ``data-dependent'' because 1) For a given input pair $(\zb_1, \zb_2)$, the spectral distributions and the kernel values $\kappa(\zb_1, \zb_2)$ depend on the input pair $(\zb_1, \zb_2)$. Thus, $\zb_1-\zb_2 = \zb_3-\zb_4$ does not necessarily imply $\kappa(\zb_1, \zb_2)= \kappa(\zb_3,\zb_4)$. One example where this phenomenon appears is symmetric positive definite kernel defined on a Riemannian manifold whose value depends on the  geodesic distance between two points rather than the Euclidean distance.
2) The marginal distribution $p(\omegab)$ depends on specific datasets, which could induce different formulas on different datasets. 

\paragraph{Constructing a data-dependent KernelNet}
To construct a data-dependent spectral distribution, we extend \eqref{eq:kernel_part_1} to the following generating process: 
\begin{equation*}
    \omegab_{\psib_2, \zb_1, \zb_2} = g_{\psib_2}(\epsilonb, \zb_1, \zb_2), ~~ \text{ where } \epsilonb\sim \mathcal{U}[\mathbf{-1}, \mathbf{1}]. 
\end{equation*} 
Note that such an implicit construction requires multiple noise samples to approximate the distribution of $\omegab_{\psib_2, \zb_1, \zb_2}$ for every $(\zb_1, \zb_2)$ pair, which could be time and space consuming when the mini-batch sizes are large. To avoid this issue, we utilize the reparameterization trick in the data-dependent sampling process, which also plays an important role later in the proof of Theorem \ref{thm:kernel_part_2_postive_definite}. 
Specifically, given an input pair $(\zb_1, \zb_2)$, we define a data-dependent sampling process as follows:
{\small
\begin{align}\label{eq:kernel_part_2}
    &\omegab_{\psib_2, \zb_1, \zb_2} = \mub_{\psib_2,\zb_1,\zb_2} + \epsilonb \odot \sigmab_{\psib_2, \zb_1, \zb_2}, \text{ where } \epsilonb \sim  \mathcal{U}[\mathbf{-1}, \mathbf{1}]\\
    &\mub_{\psib_2, \zb_1, \zb_2}=t_{\psib_2^1}(\zb_1) + t_{\psib_2^1}(\zb_2),  \sigmab_{\psib_2, \zb_1, \zb_2}=\text{exp}(t_{\psib_2^2}(\zb_1) +  t_{\psib_2^2}(\zb_2)),  \nonumber
\end{align}}
where $t_{\psib_2^1}$ and $t_{\psib_2^2}$ are two neural networks parameterized by $\psib_2^1$ and $\psib_2^2$; $\odot$ denotes the element-wise multiplication, exp denotes element-wise exponential. Since $\omegab_{\psib_2, \zb_1, \zb_2}$ depends on input pair $(\zb_1, \zb_2)$, its probability distribution is data-dependent.  There are also other ways to construct data-dependent spectral distribution, which are provided in the Appendix along with some experimental comparisons.
An important theoretical problem is to guarantee the positive definiteness of the data-dependent kernel induced by \eqref{eq:kernel_part_2}, which is shown in Theorem~\ref{thm:kernel_part_2_postive_definite}.
\begin{theorem}\label{thm:kernel_part_2_postive_definite}
    Let $\omegab$ be sampled following \eqref{eq:kernel_part_2}. The resulting kernel $\kappa_{\psib_2}(\zb_1, \zb_2) = \int _{R^d}p_{\psib_2}(\omegab \vert \zb_1, \zb_2)e^{j\omegab^\intercal(\zb_1-\zb_2)}d\omegab $ is positive definite.
\end{theorem}
\begin{wrapfigure}{R}{0.3\textwidth}
	\centering
	\includegraphics[width=0.3\textwidth]{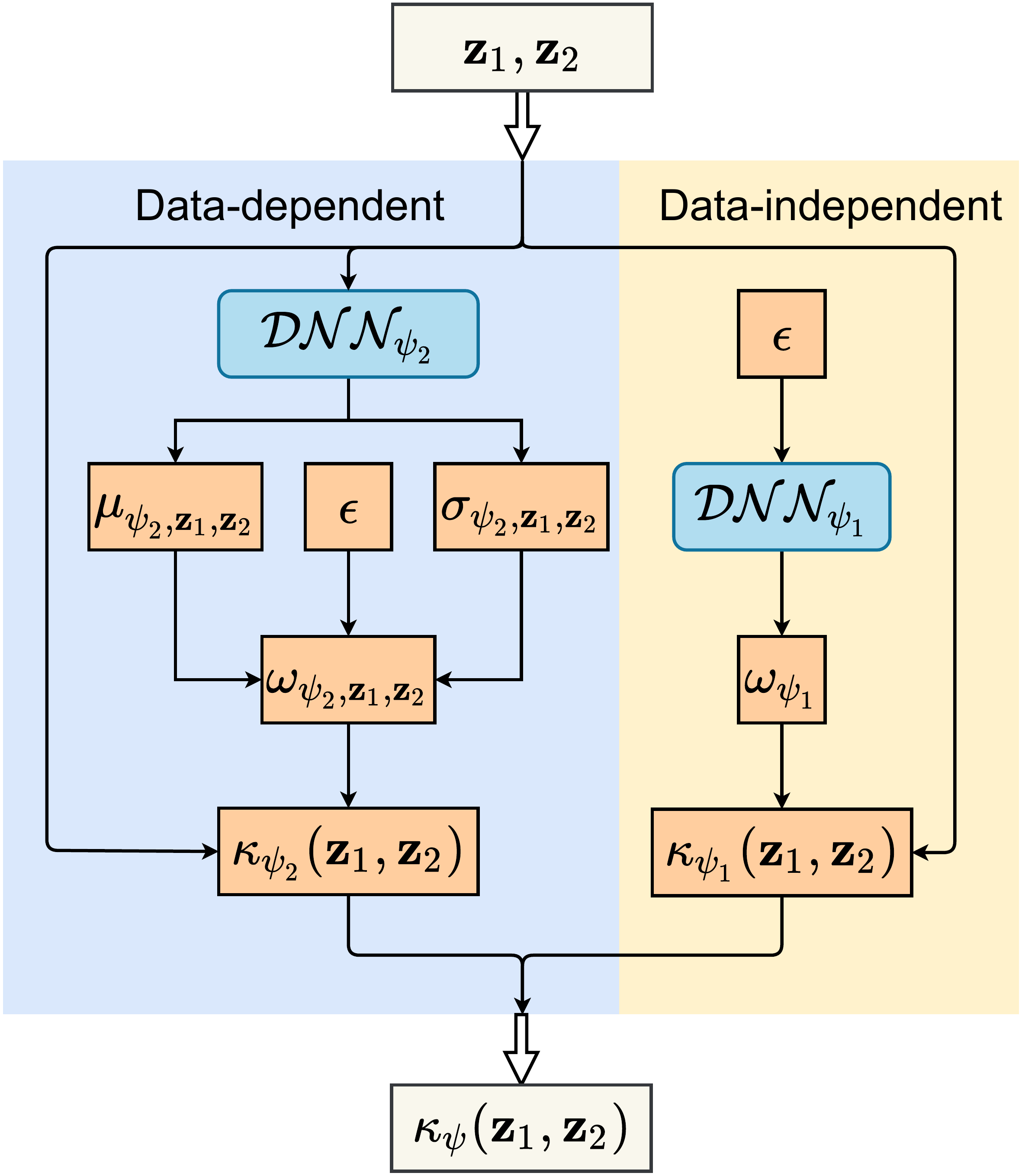}
	\vspace{-0.3cm}
    \caption{Structure of the proposed KernelNet, where $\epsilonb\sim \mathcal{U}[\mathbf{-1}, \mathbf{1}]$. }
	\vspace{-1.5cm}
    \label{fig:kernel_net}
\end{wrapfigure}
Based on the above theorem and discussions, we propose to construct our KernelNet with two components: a data-independent component and a data-dependent component. Specifically, KernelNet is constructed as follows:
\begin{flalign}\label{eq:kernelp}
    &\kappa_{\psib}(\zb_1, \zb_2) \triangleq \kappa_{\psib_1}(\zb_1, \zb_2) + \kappa_{\psib_2}(\zb_1, \zb_2), \text{ where }\\
    &\kappa_{\psib_1}(\zb_1, \zb_2) = \mathbb{E}_{\omegab_{\psib_1}, b}\left[2  \cos(\omegab_{\psib_1}^\intercal\zb_1 + b)\cos(\omegab_{\psib_1}^\intercal\zb_2 + b)\right],\nonumber\\
    &\kappa_{\psib_2}(\zb_1, \zb_2) = \mathbb{E}_{\omegab_{\psib_{2},\z_1, \z_2}, b}\left[ 2 \cos(\omegab_{\psib_2, \z_1, \z_2}^\intercal\zb_1 + b) \cos(\omegab_{\psib_2, \z_1, \z_2}^\intercal\zb_2 + b)\right]. \nonumber
\end{flalign}
$\omegab_{\psib_1}$ and $\omegab_{\psib_{2}}$ follow \eqref{eq:kernel_part_1} and \eqref{eq:kernel_part_2}, respectively, and $b \sim \mathcal{U}\left[0, 2\pi\right]$. Such a decomposition guarantees implicity and data-dependency of the kernel, and thus will not lose generalization. The network structure is illustrated in Figure~\ref{fig:kernel_net}. It is worth noting that our Kernel reduces to the one in \cite{li2019implicit} when removing the data-dependent component. We will show in experiments that the data-dependent component indeed plays an important role, and lead to performance improvement in different tasks.

In implementation, expectations are approximated by samples, e.g. for $\kappa_{\psib_1}$ in \eqref{eq:kernelp}:
\begin{align*}
    \kappa_{\psib_1}(\zb_1, \zb_2) \approx \dfrac{2}{N}\sum _{i=1}^N \cos(\omegab_{\psib_1i}^\intercal\zb_1 + b_i)\cos(\omegab_{\psib_1i}^\intercal\zb_2 + b_i), 
\end{align*}
where all $\omegab_{\psib_1i}$'s are samples from the spectral distributions $p(\omegab_{\psib_1})$ through \eqref{eq:kernel_part_1}. In addition, $b_i$'s are drawn from $p(b) = \mathcal{U}[0, 2\pi]$. 

Since the construction is implicit with no stochastic intermediate nodes, standard back-propagation can be applied for efficient end-to-end training. Lemma~\ref{lem:sumkernel} below indicates that the summation of two kernels is still a kernel, guaranteeing that the output of the KernelNet \eqref{eq:kernelp} is still a legitimate kernel. 

\begin{lemma}\cite{10.5555/975545}\label{lem:sumkernel}
    Let $\kappa_1(\zb_1, \zb_2)$ and $\kappa_2(\zb_1, \zb_2)$ be two valid kernels over $R^d\times R^d$. Then, $\kappa'(\zb_1, \zb_2)\triangleq \kappa_1(\zb_1, \zb_2) + \kappa_2(\zb_1, \zb_2)$ is also a valid kernel.
\end{lemma}


\subsection{KernelNet for MMD-GAN}
In this section, we incorporate the proposed KernelNet into learning the MMD-GAN model. 
We seek to develop an algorithm to jointly optimize both the KernelNet and the MMD-GAN model. A straightforward way is to replace the standard kernel in MMD-GAN with the proposed data-dependent kernel \eqref{eq:kernelp}. However, as the standard MMD-GAN fails to satisfy  {\em continuity in weak topology} \cite{DBLP:conf/nips/ArbelSBG18}, it is unclear whether the variant with our KernelNet would satisfy the property. To this end, we first define {\em continuity in weak topology}.

\begin{definition}[Continuity in weak topology \cite{DBLP:conf/nips/ArbelSBG18}]
$\textup{MMD}_{k}(\mathbb{Q}, \mathbb{P})$ is said to endow  continuity in weak topology if ~$\mathbb{Q} \xrightarrow[]{D} \mathbb{P}$ implies $\textup{MMD}_{k}(\mathbb{Q}, \mathbb{P})\xrightarrow{} 0$, where $\xrightarrow[]{D}$ means convergence in distribution.
\end{definition}

Continuity in weak topology in MMD-GAN is important because it makes a loss provide better signal to the generator as $\mathbb{Q}$ approaches $\mathbb{P}$, without suffering from sudden jump as in the Jensen-Shannon (JS) divergence or KL divergence (e.g. Example 1 in \cite{ArjovskyCB:arxiv17}). MMD in MMD-GAN without constraint may not be continuous in weak topology, leading to training instability and poor performance. To deal with this problem, a number of methods have been introduced ({\it e.g.}, weight-clipping \cite{DBLP:conf/nips/LiCCYP17}, gradient penalty \cite{GulrajaniAADC:NIPS17}, spectral normalization \cite{DBLP:conf/iclr/MiyatoKKY18}, and scaled objective (SMMD-GAN) \cite{DBLP:conf/nips/ArbelSBG18}), which can alleviate this issue to certain extent. To provide a theoretically guaranteed solution,
we show that  
adopting our KernelNet in MMD-GAN can lead to continuity in weak topology easily.

\begin{theorem}\label{pro:MMD_kernel}
By parameterizing the kernel with our KernelNet $\kappa_{\psib}(\zb_1, \zb_2) = \kappa_{\psib_1}(\zb_1, \zb_2) + \kappa_{\psib_2}(\zb_1, \zb_2)$, $\textup{MMD}_{\phib, \psib}(\mathbb{P}, \mathbb{Q})$ is continuous in  weak topology if the following are satisfied:
\begin{flalign}
   &\mathbb{E}_{\omegab_{\psib_1}}\left[\Vert \omegab_{\psib_1}\Vert^2\right] < \infty, ~    \mathbb{E}_{\omegab_{\psib_2, \zb_1, \zb_2}}\left[\Vert \omegab_{\psib_2, \zb_1, \zb_2}\Vert^2\right] <\infty, ~ \nonumber\\ 
   &\mathbb{E}_{\omegab_{\psib_2, \zb_1, \zb_2}}\left[\Vert \dfrac{\partial \omegab_{\psib_2, \zb_1, \zb_2}}{\partial \zb_1} - \dfrac{\partial \omegab_{\psib_2, \zb_1, \zb_2}}{\partial \zb_2}\Vert_{\mathcal{F}}\right] < \infty,  \sup _{\phib \in \Phib}\Vert h_{\phib} \Vert_{Lip} < \infty,     
\end{flalign}\par \vspace{-0.3cm}
where $\Vert \cdot \Vert_{\mathcal{F}}$ denotes the Frobenius norm of a matrix, $h_{\phib}$ is the injective function in MMD-GAN, {\it i.e.}, $\zb=h_{\phib}(\xb)$, and $\Vert h_{\phib} \Vert_{Lip}$ denotes its Lipschitz constant.
\end{theorem}

Based on Theorem~\ref{pro:MMD_kernel}, we propose several variants of the MMD-GAN model, respectively corresponding to the MMD-GAN \cite{DBLP:conf/nips/LiCCYP17}, SMMD-GAN \cite{DBLP:conf/nips/ArbelSBG18} and MMD-GAN with repulsive loss (denoted as Rep-GAN) \cite{wang2018improving}, by incorporating the conditions in Theorem~\ref{pro:MMD_kernel} into the objective functions. 

\paragraph{MMD-GAN with the KernelNet}
By adopting spectral normalization and the method of Lagrange multipliers to regularize the conditions in Theorem~\ref{pro:MMD_kernel}, we propose SN-MMD-GAN-DK. Note that $\sup _{\phib \in \Phib}\Vert h_{\phib} \Vert_{\text{Lip}} < \infty$ is satisfied because of the spectral normalization operation, which normalizes the weight matrix during the training process. The objective of the generator parameterized by $\thetab$ and discriminator parameterized by $(\phib, \psib)$ are defined as:
\begin{align}\label{eq:mmd_dk}
    &\min_{\thetab} \textup{MMD}^2_{\phib, \psib}(\mathbb{P}, \mathbb{Q}) + \alpha_1 \Omega(\thetab, \phib, \psib)~, \text{ and } 
    \min _{\phib, \psib} -\textup{MMD}^2_{\phib, \psib}(\mathbb{P}, \mathbb{Q}) + \alpha_2 \Omega(\thetab, \phib, \psib)~, \text{ where } \\
    &\Omega(\thetab, \phib, \psib)  \triangleq  \mathbb{E}_{\omegab_{\psib_1}}\left[\Vert \omegab_{\psib_1}\Vert^2\right]
     + \mathbb{E}_{\omegab_{\psib_2, \zb_1, \zb_2}}[\Vert \omegab_{\psib_2, \zb_1, \zb_2}\Vert^2+ \Vert \dfrac{\partial \omegab_{\psib_2, \zb_1, \zb_2}}{\partial \zb_1} - \dfrac{\partial \omegab_{\psib_2, \zb_1, \zb_2}}{\partial \zb_2}\Vert_{\mathcal{F}}]. \label{eq:omega_regularizer}
\end{align}
\paragraph{Scaled MMD-GAN with the KernelNet}
Similarly, based on the SMMD-GAN model \cite{DBLP:conf/nips/ArbelSBG18}, we propose our variant SN-SMMD-GAN-DK by incorporating the conditions in Theorem~\ref{pro:MMD_kernel} into the SMMD framework.
\begin{proposition}\label{pro:SMMD_kernel}
With the proposed data-dependent KernelNet \eqref{eq:kernelp}, the SMMD-DK framework  can be formulated and simplified as:
\[
\textup{SMMD-DK}_{\phib, \psib,\lambda}(\mathbb{P}, \mathbb{Q}):= \sigma_{\phib, \psib,\lambda}\textup{MMD}_{\phib, \psib}(\mathbb{P}, \mathbb{Q}),~~\text{where }\zb = h_{\phib}(\xb) \text{ and}
\]
{\small
\begin{align*}
    \sigma_{\phib, \psib, \lambda}:=\bigg\{&\lambda + 1 + \mathbb{E}_{\xb \sim \mathbb{P}}\{\mathbb{E}_{\omegab_{\psib_2, \zb, \zb}}[\Vert\omegab_{\psib_2, \zb, \zb}\Vert^2]\Vert\nabla h_{\phib}(\xb)\Vert _{\mathcal{F}}\} +\mathbb{E}_{\xb \sim \mathbb{P}}\{\mathbb{E}_{\omegab_{\psib_1}}[\Vert\omegab_{\psib_1}\Vert^2]\Vert\nabla h_{\phib}(\xb)\Vert _{\mathcal{F}}\}\bigg\}^{-1/2}.
\end{align*}
    }
\end{proposition}

Consequently, by incorporating the conditions in Theorem~\ref{pro:MMD_kernel}, the objectives for generator and discriminator in SN-SMMD-GAN-DK are defined as:
{\small\begin{align}\label{eq:smmd_dk}
    \min_{\thetab} \textup{SMMD-DK}^2_{\phib, \psib,\lambda}(\mathbb{P}, \mathbb{Q}) + \alpha_1 \Omega(\thetab, \phib, \psib)~, \text{ and }\min _{\phib, \psib} -\textup{SMMD-DK}^2_{\phib, \psib,\lambda}(\mathbb{P}, \mathbb{Q}) + \alpha_2 \Omega(\thetab, \phib, \psib)~, 
\end{align}}
where $\Omega(\thetab, \phib, \psib)$ is defined as in  \eqref{eq:omega_regularizer}. 
In practice, we choose $\lambda$ and scale the original SMMD-DK obejective so that the ``$\textup{SMMD-DK}^2$'' in \eqref{eq:smmd_dk} is replaced by the following:
\begin{align}\label{eq:real_smmd_obj}
    &\widehat{\textup{SMMD-DK}}^2_{\psib, \phib,\zeta}(\mathbb{P}, \mathbb{Q}) = \delta_{\phib,\psib,\zeta}\textup{MMD}^2_{k_{\psib}}(\mathbb{P}, \mathbb{Q}) \text{, where} \\
    &\delta_{\phib, \psib, \zeta} :=\bigg\{1 + \zeta\mathbb{E}_{\xb \sim \mathbb{P}}\{\mathbb{E}_{\omegab_{\psib_2, \zb, \zb}}[\Vert\omegab_{\psib_2, \zb, \zb}\Vert^2]\Vert\nabla h_{\phib}(\xb)\Vert _{\mathcal{F}}\}   +\mathbb{E}_{\xb \sim \mathbb{P}}\{\mathbb{E}_{\omegab_{\psib_1}}[\Vert\omegab_{\psib_1}\Vert^2]\Vert\nabla h_{\phib}(\xb)\Vert _{\mathcal{F}}\}\bigg\}^{-1} \nonumber
\end{align}

\paragraph{Repulsive loss with the KernelNet}
By incorporating KernelNet into the repulsive loss, we further propose Rep-GAN-DK. According to Theorem \ref{pro:MMD_kernel}, the objective functions for generator and discriminator in Rep-GAN-DK are defined as:
\begin{align}
    &\min_{\thetab} \textup{MMD}^2_{\phib, \psib}(\mathbb{P}, \mathbb{Q}) + \alpha_1 \Omega(\thetab, \phib, \psib)~, \text{and} \min _{\phib, \psib} L_{\eta, \phib, \psib} + \alpha_2 \Omega(\thetab, \phib, \psib), \text{ where} \label{eq:mmd_rep_dk}\\
    &L_{\eta, \phib, \psib} = \eta\mathbb{E}_{\xb, \xb' \sim \mathbb{P}}\left[ k_{\phib, \psib}(\xb, \xb')\right] - \mathbb{E}_{\yb,\yb' \sim \mathbb{Q}}[k_{\phib, \psib}(\yb, \yb')]  - (\eta-1)\mathbb{E}_{\xb \sim \mathbb{P}, \yb \sim \mathbb{Q}}[k_{\phib, \psib}(\xb, \yb)], \label{eq:rep_loss}
\end{align}
$\Omega(\thetab, \phib, \psib)$ is defined as \eqref{eq:omega_regularizer}, $\eta$ is a hyper-parameter. One can find that when $\eta = -1$, \eqref{eq:mmd_rep_dk} will reduce to the standard MMD case \eqref{eq:mmd_dk}.

\subsection{KernelNet for Implicit Info-VAE}
In this section, we describe how to incorporate our KernelNet into the Info-VAE framework. First, to increase the power of Info-VAE, we adopt an implicit encoder setting. That is, instead of adopting a particular posterior distribution family such as Gaussian for the encoder, we construct a complex implicit distribution by adding random noise at each layer of the encoder (including input data) and removing the reparameterization trick. 

One problem with such a method is the need of evaluating the density of the implicit encoder distribution for model training, as seen in the objective \eqref{eq:infovae_obj}. To deal with this issue, we adopt the Stein gradient estimator (SGE) \cite{LiT:ICLR18} to approximate the gradient of the log-density. 

Another problem is the difficulty of computing mutual information (MI). MI between two distributions is tractable only in certain situations, e.g. both distributions are Gaussian. In the implicit VAE setting, one has to design some non-trivial methods to deal with the intractability of the mutual information. In our work, we propose to replacing the mutual information with MMD. The logic is quite straightforward because both MMD and MI can be reconsidered as distance measures of two distributions. MMD is much easier to be dealt with, because it can be computed based on samples regardless of how complex the distributions are. Consequently, we apply our proposed KernelNet to the computation of MMD, which leads to the following objective:
{\small
\begin{align*}
    \max _{\phib, \thetab, \psib} -\lambda \textup{KL}[q_{\phib}(\zb)\Vert p(\zb)] - \mathbb{E}_{q_{\phib}(\zb)}\{\textup{KL}[q_{\phib}(\xb \vert \zb)\Vert p_{\thetab}(\xb \vert \zb)]\} + \alpha\mathbb{E}_{\q(\xb)}\{\text{MMD}_{\psib}[q_{\phib}(\zb \vert \xb), q_{\phib}(\zb)]\}~,
\end{align*}
}
with hyper-parameter $\lambda$ and $\alpha$. The objective can be further reformulated as:
{\small
\begin{flalign}\label{eq:infovae_obj_mmd}
    \max _{\phib, \thetab, \psib} &\ \mathbb{E}_{q(\xb)}\mathbb{E}_{q_{\phib}(\zb\vert \xb)}[\log p_{\thetab}(\xb \vert \zb)] -\mathbb{E}_{q(\xb)}\{\text{KL}[q_{\phib}(\zb\vert \xb) \Vert p(\zb)]\} \nonumber\\
    & - \mathbb{E}_{q(\xb)}(\log q(\xb)) -(\lambda-1)\text{KL}[q_{\phib}(\zb) \Vert p(\zb)] + \alpha\mathbb{E}_{\q(\xb)}\{\text{MMD}_{\psib}[q_{\phib}(\zb \vert \xb), q_{\phib}(\zb)]\}~.
\end{flalign}
}
Note that $\mathbb{E}_{q(\xb)}(\log q(\xb))$ is independent of the model and can be discarded in optimization. Our proposed model is very general: when $\lambda = 1$ and $\alpha=0$, \eqref{eq:infovae_obj_mmd} reduces to the objective of vanilla VAE.


\section{Experiments}
We conduct experiments to test the performance of our proposed KernelNet applied to variants of MMD-GAN and implicit VAE, and compare them with related methods, including MMD and non-MMD based GANs, semi-implicit and implicit VAE models. Our experiments are implemented using Tensorflow on a Nvidia Titan Xp GPU, all the code will be available online.
\subsection{MMD-GAN}
We evaluated our MMD-GAN variants on four datasets: CIFAR-10, STL-10, ImageNet and CelebA. Following \cite{DBLP:conf/nips/ArbelSBG18} and \cite{wang2018improving}, we scale training images from these datasets to the resolution of $32\times 32$, $48 \times 48$, $64 \times 64$ and $160 \times 160$ respectively.

\begin{wraptable}{R}{0.4\textwidth}
    \vspace{-1.0cm}
    \caption{IKL vs. KernelNet}
    \label{tab:comparison_ikl_dk}  
    \begin{center}
        \begin{sc}
        \begin{adjustbox}{scale=0.65}
            \begin{tabular}{lcc}
            \toprule
            &\multicolumn{2}{c}{CIFAR-10}\\
            & FID $(\downarrow)$ & IS $(\uparrow)$\\
            \midrule
             SN-MMD-GAN&$31.5\pm 0.2$&$6.9\pm 0.1$\\
             
             SN-MMD-GAN-IKL&$30.4\pm0.1$&$6.9\pm0.1$\\ 
             
             SN-MMD-GAN-DK (ours) &$\mathbf{27.7 \pm 0.1}$&$\mathbf{7.2 \pm 0.1}$\\
            \midrule
             SN-SMMD-GAN&$25.0\pm 0.3$&$7.3\pm 0.1$\\

             SN-SMMD-GAN-IKL&$26.4\pm 0.1$&$7.3\pm 0.1$\\

             SN-SMMD-GAN-DK (ours)&$\mathbf{24.3 \pm 0.1}$&$\mathbf{7.4 \pm 0.1}$\\
             \midrule
             Rep-GAN&$16.7$&$8.0 $\\

             Rep-GAN-IKL&$16.3 \pm 0.1$&$8.0 \pm 0.1$\\

             Rep-GAN-DK (ours)&$\mathbf{14.6 \pm 0.1}$&$\mathbf{8.2 \pm 0.1}$\\
            \bottomrule
            \end{tabular}  
        \end{adjustbox}
        \end{sc}   
    \end{center}
    \vskip -0.1in
\end{wraptable}
We compare our models with WGAN-GP \cite{GulrajaniAADC:NIPS17}, MMD-GAN \cite{DBLP:conf/nips/LiCCYP17}, SN-GAN \cite{DBLP:conf/iclr/MiyatoKKY18}, SMMD-GAN, SN-SMMD-GAN \cite{DBLP:conf/nips/ArbelSBG18}, Rep-GAN \cite{wang2018improving}, CR-GAN \cite{zhang2020consistency} and report the standard Fréchet Inception Distance (FID) \cite{DBLP:journals/corr/HeuselRUNKH17} and Inception Score (IS) \cite{SalimansGZCRC:NIPS16}. Due to the limitation of space, we leave detailed experimental settings in Appendix \ref{app:setting}.

To illustrate the effectiveness of our data-dependent component, we first compare with models using kernels without the data-dependent component, which is the same as the IKL method proposed by \cite{li2019implicit}. These models are denoted as SN-MMD-GAN-IKL, SN-SMMD-GAN-IKL and Rep-GAN-IKL. The results are reported in Table \ref{tab:comparison_ikl_dk}. As we can see, our KernelNet-based models obtain best results, showing the importance of data-dependent component. 
\begin{table*}[tb!]
    \caption{Results of image generation.}
    \label{tab:results_image_generation}  
    \vspace{-0.1in}
    \begin{center}
        \begin{sc}
        \begin{adjustbox}{scale=0.7}
            \begin{tabular}{lcccccccc}
            \toprule
            &\multicolumn{2}{c}{CIFAR-10} & \multicolumn{2}{c}{STL-10} &\multicolumn{2}{c}{CelebA} & \multicolumn{2}{c}{ImageNet}\\
            & FID $(\downarrow)$ & IS $(\uparrow)$ & FID $(\downarrow)$ & IS $(\uparrow)$ & FID $(\downarrow)$ & IS $(\uparrow)$ & FID $(\downarrow)$ & IS $(\uparrow)$\\
            \midrule
            WGAN-GP & $31.1 \pm 0.2$ & $6.9 \pm 0.2$ & $55.1$ & $8.4 \pm 0.1$ & $29.2 \pm 0.2$ & $2.7 \pm 0.1$ & $65.7 \pm 0.3$ & $7.5 \pm 0.1$\\
            SN-GAN & $25.5$ & $7.6 \pm 0.1$ & $43.2$ & $8.8 \pm 0.1$ & $22.6 \pm 0.1$ & $2.7 \pm 0.1$ & $47.5 \pm 0.1$ & $11.2 \pm 0.1$\\
            SMMD-GAN & $31.5 \pm 0.4$ & $7.0 \pm 0.1$& $43.7 \pm 0.2$ & $8.4 \pm 0.1$ & $18.4 \pm 0.2$ & $2.7 \pm 0.1$ & $38.4 \pm 0.3$ & $10.7 \pm 0.2$\\
            SN-SMMD-GAN & $25.0 \pm 0.3$ & $7.3 \pm 0.1$ & $40.6 \pm 0.1$ & $8.5 \pm 0.1$ & $12.4 \pm 0.2$ & $2.8 \pm 0.1$ & $36.6 \pm 0.2$ & $10.9 \pm 0.1$\\
            SN-SMMD-GAN-DK (ours) & $24.3\pm 0.1$ & $7.4 \pm 0.1$ & $40.0 \pm 0.1 $ & $8.5 \pm 0.1$ & $\mathbf{11.3 \pm 0.1}$ & $\mathbf{2.9 \pm 0.1}$ & $35.7 \pm 0.3$ & $11.2\pm 0.2$\\
            CR-GAN & $18.7$ & $7.9$& -- & -- & -- & -- & -- & --\\
            Rep-GAN & $16.7$ & $8.0$ & $36.7$ & $\mathbf{9.4}$ & $16.8 \pm 0.1 $ & $2.9 \pm 0.1$ & $31.0 \pm 0.1$ & $11.5 \pm 0.1$\\
            Rep-GAN-DK (ours) & $\mathbf{14.6 \pm 0.1} $ & $\mathbf{8.2 \pm 0.1}$ & $\mathbf{33.4\pm 0.1}$ &$9.3 \pm 0.1$& $16.1 \pm 0.1$ & $2.9 \pm 0.1$ & $\mathbf{30.1 \pm 0.1}$ & $\mathbf{11.7 \pm 0.1}$\\
            \bottomrule
            \end{tabular}  
        \end{adjustbox}
        \end{sc}
    \end{center}
    \vskip -0.1in
\end{table*}

In addition, the results on more models with different datasets are summarized in the Table \ref{tab:results_image_generation}, with the generated examples shown in Figure \ref{fig:generated_img_rep_dk}. Some results are taken from the corresponding papers when we can not reproduce their results, thus may not have standard deviations. We can see that our proposed method achieves competitive results on all the datasets, consistently improve different variants of MMD-GANs.
\begin{table*}[ht!]
    \vspace{-0.1in}
    \caption{Negative log-likelihood on the binarized MNIST dataset.}\label{tab:mnist_nll}
    \vskip 0.1in
    \begin{center}
    \begin{sc}
        \begin{adjustbox}{scale=0.7,center}
            \begin{tabular}{lccccccccc}
            \toprule
             Model&VAE&Stein-VAE&Spectral&SIVI&Info-VAE&Info-IVAE&Info-IVAE-RBF&Info-IVAE-IKL&Info-IVAE-DK \\
             \midrule
             NLL~$\downarrow$ &90.32&88.85&89.67&89.03&88.89&89.79&88.24&88.21&\textbf{88.16}\\
             \bottomrule
            \end{tabular}
        \end{adjustbox}
    \end{sc}
    \end{center}
    \vskip -0.1in
\end{table*}

\begin{figure}[tb!]
    \vspace{-0.1in}
    \centering
    \subfigure[CIFAR-10($32 \times 32$)]{\includegraphics[width=0.23\linewidth]{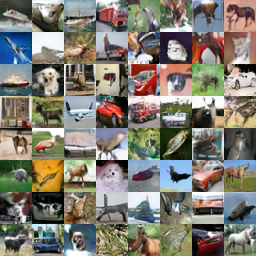}
    \label{fig:cifar_32}}
    \subfigure[STL-10($48 \times 48$)]{\includegraphics[width=0.23\linewidth]{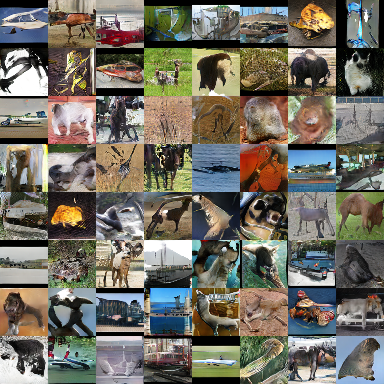}
    \label{fig:stl_32}}
    \subfigure[ImageNet ($64 \times 64$)]{\includegraphics[width=0.23\linewidth]{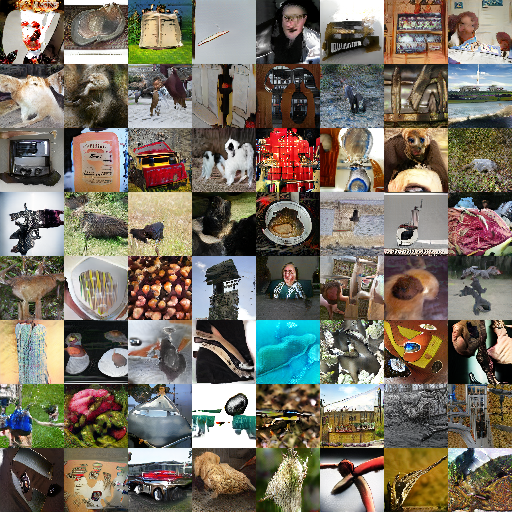}
    \label{fig:imagenet_64}}
    \subfigure[CelebA ($160 \times 160$)]{\includegraphics[width=0.23\linewidth]{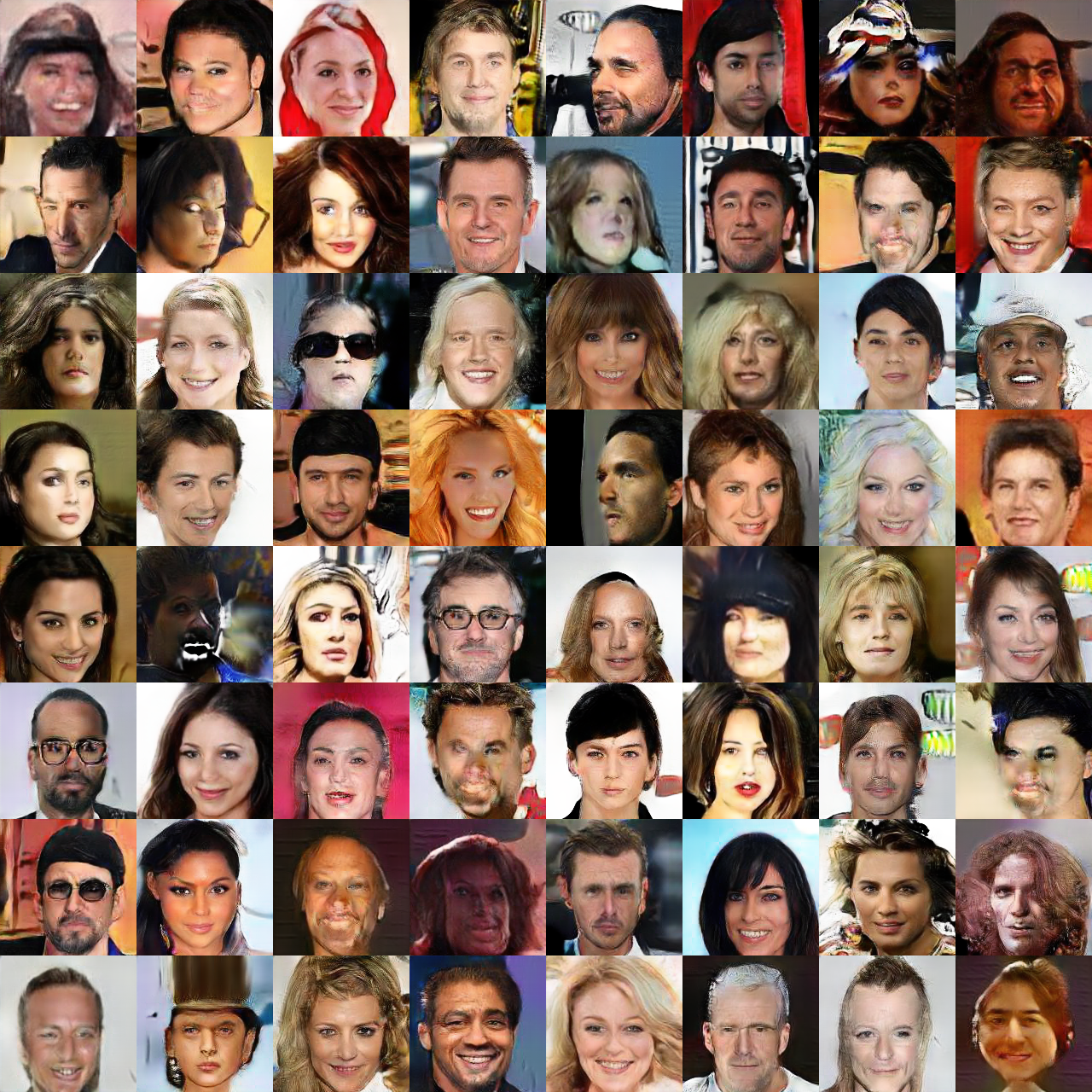}
    \label{fig:celeba_160}}
    \caption{Generated images of Rep-GAN-DK.}\label{fig:generated_img_rep_dk}
    \vspace{-0.1in}
\end{figure}

\subsection{Implicit VAE}
Due to the limitation of space, we only show the results of our Implicit Info-VAE model on the MNIST dataset \cite{DBLP:conf/icml/SalakhutdinovM08} here, and leave other experiments along with detailed settings in the Appendix \ref{app:results}. In this experiment, we use a network with 1 fully-connected hidden layer for both encoder and decoder, whose hidden units are set to 400. 
Bernoulli noises are injected into the encoder by using dropout with a dropout rate of 0.3. At every step, 512 random features from the spectral distribution are sampled. 

For fair evaluation, we follow \cite{DBLP:journals/corr/WuBSG16} and use Annealed Importance Sampling (AIS) to approximate the negative log-likelihood (NLL). 10 independent AIS chains are used, each of which have 1000 intermediate distributions. The final results are computed using 5000 random sampled test data. The results are shown in Table \ref{tab:mnist_nll}, and some reconstructed and generated images and t-sne visualization are provided in the Appendix \ref{app:results}. We compare with related models including: VAE (vanilla VAE from \cite{KingmaW:ICLR14}), Stein-VAE (amortized SVGD from \cite{FengWL:UAI17}), SIVI (Semi-Implicit VAE from \cite{DBLP:conf/icml/YinZ18}), Spectral (implicit VAE with spectral method for gradient estimation from \cite{shi2018spectral}) and Info-VAE \cite{DBLP:journals/corr/ZhaoSE17b}. Note that some models have also reported scores related to NLL in their original paper under different settings, which are not directly comparable to ours. For fair comparisons, we rerun all the models with the same model structure. We denote our Implicit Info-VAE with Stein gradient estimator with objective \eqref{eq:infovae_obj} as Info-IVAE. The models with objective \eqref{eq:infovae_obj_mmd} are denoted as Info-IVAE-RBF, Info-IVAE-IKL and Info-IVAE-DK, where the MMD regularizers are computed by RBF kernel, IKL and KernelNet respectively. Our model obtains the best NLL score among all the models.

\section{Conclusion}
We propose KernelNet, a novel way of parameterizing learnable data-dependent kernels using implicit spectral distributions parameterized by DNNs. We prove the positive definiteness of our KernelNet, and present how the proposed KernelNet can be applied to deep generative models, including several variants of MMD-GAN and Info-VAE, along with some theoretical analysis. Experiments show that the proposed  KernelNet leads to performance improvement over related models, demonstrating the effectiveness of data-dependent kernels.
\clearpage


\bibliography{references.bib}
\bibliographystyle{unsrt}

\newpage 
\appendix
\onecolumn
\phantom{}
\section{Proof of Theorem \ref{pro:kernel_cos}}\label{app:kernel_cos}
\textbf{Theorem \ref{pro:kernel_cos}} 
\textit{
    Let $\omegab$ be drawn from some spectral distribution $p(\omegab)$, and $b$ be drawn uniformly from $[0, 2\pi]$. The real-valued kernel in \eqref{eq:kernel} can be reformulated into the following two forms:
    \begin{flalign}
        \kappa(\zb_1, \zb_2) &= \mathbb{E}_{\omegab \sim p(\omegab)}\{\cos\left[\omegab^\intercal(\zb_1-\zb_2)\right]\}\label{eq:kernel_cos_0}\\
        &= \mathbb{E}_{\omegab, b}\left[2\cos(\omegab^\intercal\zb_1 +b)\cos(\omegab^\intercal\zb_2+b) \label{eq:kernel_cos} \right]
    \end{flalign}
    where $\omegab \sim p(\omegab), b\sim \mathcal{U}[0, 2\pi]$.
    }
\begin{proof}
By Euler's formula, we have:
\begin{equation*}
    \kappa(\zb_1, \zb_2) = \int _{R^d}p(\omegab)e^{j\omegab^\intercal(\zb_1-\zb_2)}d\omegab = \mathbb{E}_{\omegab}\{\cos\left[\omegab^\intercal(\zb_1 -\zb_2)\right] +j \sin \left[\omegab^\intercal(\zb_1 -\zb_2)\right]\}
\end{equation*}
For real-valued kernel, we remove the imaginary part, we have:
\begin{equation*}
    \kappa(\zb_1, \zb_2) = \mathbb{E}_{\omegab}\{\cos\left[\omegab^\intercal(\zb_1 -\zb_2)\right]\}
\end{equation*}
Now we show $\mathbb{E}_{\omegab}\{\cos\left[\omegab^\intercal(\zb_1 -\zb_2)\right]\}=2\mathbb{E}_{\omegab, b}\left[\cos(\omegab^\intercal\zb_1 +b)\cos(\omegab^\intercal\zb_2 +b)\right]$, where b follows a uniform distribution $\mathcal{U}[0, 2\pi]$:
\begin{flalign*}
    &2\mathbb{E}_{\omegab, b}\left[\cos(\omegab^\intercal\zb_1 +b)\cos(\omegab^\intercal\zb_2 +b)\right]\\
    =&2\mathbb{E}_{\omegab}\mathbb{E}_{b}\{\left[\cos(\omegab^\intercal\zb_1) \cos b - \sin (\omegab^\intercal\zb_1)\sin b\right]\left[\cos(\omegab^\intercal\zb_2) \cos b - \sin (\omegab^\intercal\zb_2)\sin b\right]\}\\
    =&2\mathbb{E}_{\omegab}\mathbb{E}_{b}(\cos \omegab^\intercal\zb_1 \cos \omegab^\intercal\zb_2 \cos ^2b - \sin \omegab^\intercal\zb_1 \cos \omegab^\intercal\zb_2 \sin b\cos b \\&\quad - \cos \omegab^\intercal \zb_1 \sin \omegab^\intercal\zb_2 \sin b\cos b + \sin \omegab^\intercal \zb_1 \sin \omegab^\intercal \zb_2 \sin ^2b)\\
    =&2\mathbb{E}_{\omegab}\mathbb{E}_{b}[\cos \omegab^\intercal\zb_1 \cos \omegab^\intercal\zb_2 \cos ^2b + \sin \omegab^\intercal \zb_1 \sin \omegab^\intercal \zb_2 \sin ^2b ] \\&\quad  -2\mathbb{E}_{\omegab}\mathbb{E}_{b} [(\sin \omegab^\intercal\zb_1 \cos \omegab^\intercal\zb_2 - \cos \omegab^\intercal \zb_1 \sin \omegab^\intercal\zb_2 )\sin ^2b]\\
    =&2\mathbb{E}_{\omegab}\mathbb{E}_{b}[\cos \omegab^\intercal\zb_1 \cos \omegab^\intercal\zb_2 \cos ^2b + \sin \omegab^\intercal \zb_1 \sin \omegab^\intercal \zb_2 \sin ^2b] \\
    =&2\mathbb{E}_{\omegab}\{\cos \omegab^\intercal\zb_1 \cos \omegab^\intercal\zb_2 \mathbb{E}_b(\cos 2b+1) + \sin \omegab^\intercal\zb_1 \sin \omegab^\intercal\zb_2\mathbb{E}_b(1-\cos 2b)\}\\
    =&\mathbb{E}_{\omegab}[(\cos \omegab^\intercal\zb_1 \cos \omegab^\intercal\zb_2 + \sin \omegab^\intercal\zb_1 \sin \omegab^\intercal\zb_2)\mathbb{E}_b(1)]\\
    =&\mathbb{E}_{\omegab}(\cos \omegab^\intercal\zb_1 \cos \omegab^\intercal\zb_2 + \sin \omegab^\intercal\zb_1 \sin \omegab^\intercal\zb_2)\\
    =&\mathbb{E}_{\omegab}\{\cos \left[\omegab^\intercal(\zb_1 -\zb_2)\right]\}
\end{flalign*}
The theorem has been proved.
\end{proof}

\section{Proof of Theorem \ref{thm:data_dependent_postive_definite}}\label{app:pd_proof}
\textbf{Theorem \ref{thm:data_dependent_postive_definite}} 
\textit{
    Let $\omegab$ be drawn from a data-dependent spectral distribution $p(\omegab\vert \zb_1, \zb_2)$. If the probability density function can be formulated as 
    $p(\omegab\vert \zb_1, \zb_2) = r(\omegab)s(\omegab, \zb_1)s(\omegab, \zb_2)$,
    where $r$ and $s$ are functions such that $r(\omegab) \geq 0$  ($\forall \omegab$) and $s(\omegab, \zb_1)s(\omegab, \zb_2) \geq 0$ ($\forall (\omegab, \zb_1, \zb_2)$). Then, the kernel $\kappa(\zb_1, \zb_2) = \int _{R^d}p(\omegab \vert \zb_1, \zb_2)e^{j\omegab^\intercal(\zb_1-\zb_2)}d\omegab $ is positive definite.
    }
\begin{proof}
For an arbitrary number n, let $c_1, ...c_n \in R$ be some constants, $\zb_1, ..., \zb_n$ be some data samples. Given $p(\omegab\vert \zb_1, \zb_2) = r(\omegab)s(\omegab, \zb_1)s(\omegab, \zb_2)$, using $i,k$ as subscripts to distinguish the elements, we have:
\begin{align*}
    &\sum_{i=1}^n\sum_{k=1}^n c_i c_k\kappa(\zb_i, \zb_k)\\ 
    =&  \sum_{i=1}^n\sum_{k=1}^n  c_i c_k\int_{R^d} p(\omegab \vert \zb_i, \zb_k) e^{j\omegab^\intercal (\zb_i -\zb_k)} d\omegab \\
    =& \sum_{i=1}^n\sum_{k=1}^n  c_i c_k\int_{R^d}  r(\omegab)s(\omegab, \zb_i)s(\omegab, \zb_k) e^{j\omegab^\intercal (\zb_i -\zb_k)} d\omegab \\
    =& \sum_{i=1}^n\sum_{k=1}^n  c_i c_k\int_{R^d}  r(\omegab)s(\omegab, \zb_i)s(\omegab, \zb_k) e^{j\omegab^\intercal\zb_i} e^{-j\omegab^\intercal\zb_k} d\omegab \\
    =&  \int_{R^d}   r(\omegab) \sum_{i=1}^n\sum_{k=1}^n c_i c_k  s(\omegab, \zb_i)s(\omegab, \zb_k) e^{j\omegab^\intercal\zb_i} e^{-j\omegab^\intercal\zb_k} d\omegab \\
    =&  \int_{R^d}   r(\omegab) \sum_{i=1}^n\sum_{k=1}^n c_i c_k^* s(\omegab, \zb_i)s^*(\omegab, \zb_k) e^{j\omegab^\intercal\zb_i} e^{-j\omegab^\intercal\zb_k} d\omegab \\
    &\text{\quad $^*$ denotes the conjugate, because $c_k, s(\omegab, \zb_k) \in R$, so $c_k^* = c_k, s^*(\omegab, \zb_k) = s(\omegab, \zb_k)$ }\\
    =& \int_{R^d}  r(\omegab) \sum_{i=1}^n c_i s(\omegab, \zb_i) e^{j\omegab^\intercal\zb_i} \sum_{k=1}^n c_k^* s^*(\omegab, \zb_k) (e^{j\omegab^\intercal\zb_k})^* d\omegab \\
    =& \int_{R^d}  r(\omegab)\sum_{i=1}^n c_i s(\omegab, \zb_i) e^{j\omegab^\intercal\zb_i}\sum_{i=1}^n \big(c_is(\omegab, \zb_i) e^{j\omegab^\intercal\zb_i}\big)^* d\omegab \text{\quad $i, k$ are nothing but subscripts}\\
    =& \int_{R^d}  r(\omegab)\big \vert \sum_{i=1}^n c_i s(\omegab, \zb_i) e^{j\omegab^\intercal\zb_i}\big \vert^2 d\omegab \\
    \geq & 0
\end{align*}
In the last inequality, $\vert \cdot \vert$ denotes the norm of complex number: $\vert x+yj\vert^2 = x^2+y^2\geq 0$. Because we assume $r(\omegab) \geq 0$ always holds for any $\omegab$, the last inequality holds.

According to the definition of positive definite kernel, $\kappa$ is positive definite if and only if for any $c_1,...,c_n \in R$ and $\zb_1, ..., \zb_n \in R^d$, $\sum_{i=1}^n\sum_{k=1}^n c_i c_k \kappa(\zb_i, \zb_k)\geq 0$ always holds. Thus our kernel is positive definite by definition.
\end{proof}

\section{Proof of Theorem \ref{thm:kernel_part_2_postive_definite}}\label{app:pd_construction}
\textbf{Theorem \ref{thm:kernel_part_2_postive_definite}} 
\textit{
    Let $\omegab$ be sampled following \eqref{eq:kernel_part_2}. Then the resulting kernel $\kappa_{\psib_2}(\zb_1, \zb_2) = \int _{R^d}p_{\psib_2}(\omegab \vert \zb_1, \zb_2)e^{j\omegab^\intercal(\zb_1-\zb_2)}d\omegab $ is positive definite.
    }
\begin{proof}
Because we have already proved Theorem \ref{thm:data_dependent_postive_definite}, we only need to show that 
\[
p_{\psib_2}(\omegab \vert \zb_1, \zb_2) = r(\omegab)s(\omegab, \zb_1)s(\omegab, \zb_2)
\]
for some function $r$, $s$ and $r(\omegab) \geq 0$ for any $\omegab$. 

Recall that
\[
    \omegab_{\psib_2, \zb_1, \zb_2} = t_{\psib_2^1}(\zb_1) + t_{\psib_2^1}(\zb_2) + \epsilonb \odot \text{exp}(t_{\psib_2^2}(\zb_1) +  t_{\psib_2^2}(\zb_2)), ~~\text{ where} \epsilonb\sim  \mathcal{U}[\mathbf{-1}, \mathbf{1}]~ 
\]
For simplicity and clearness, we use $\omegab^i$ to denote the element at $i^{th}$ dimension of $\omegab_{\psib_2, \zb_1, \zb_2}$. Because of the reparameterization and element-wise multiplication, elements on different dimensions ($\omegab^i$s) are actually independent from each other. Thus we have:
\[
p_{\psib_2}(\omegab \vert \zb_1, \zb_2) = \prod_{i=1}^d p_{\psib_2}(\omegab^i \vert \zb_1, \zb_2)
\]
By construction, our $p(\omegab^i \vert \zb_1, \zb_2)$ is a uniform distribution:
\[p(\omegab^i \vert \zb_1, \zb_2) \sim \mathcal{U}\left[ t^i_{\psib_2^1}(\zb_1) + t^i_{\psib_2^1}(\zb_2) - \text{exp}(t^i_{\psib_2^2}(\zb_1) +  t^i_{\psib_2^2}(\zb_2)),\  t^i_{\psib_2^1}(\zb_1) + t^i_{\psib_2^1}(\zb_2) + \text{exp}(t^i_{\psib_2^2}(\zb_1) +  t^i_{\psib_2^2}(\zb_2))\right]
\] 
where $t^i_{\psib_2^1}$, $t^i_{\psib_2^2}$ denote $i^{th}$ elements of $t_{\psib_2^1}$ and $t_{\psib_2^2}$ . 

Then we have:
\[
p(\omegab^i \vert \zb_1, \zb_2) = \dfrac{1}{2\text{exp}(t^i_{\psib_2^2}(\zb_1) +  t^i_{\psib_2^2}(\zb_2))} = \dfrac{1}{\sqrt{2}\text{exp}(t^i_{\psib_2^2}(\zb_1))}\dfrac{1}{\sqrt{2}\text{exp}(t^i_{\psib_2^2}(\zb_2))}
\]
Thus
\[
p_{\psib_2}(\omegab \vert \zb_1, \zb_2) = \prod_{i=1}^d p_{\psib_2}(\omegab^i \vert \zb_1, \zb_2) = r(\omegab)s_{\psib_2}(\omegab, \zb_1)s_{\psib_2}(\omegab, \zb_2)
\]
where
\[
r(\omegab) = 1 \geq 0,\ s_{\psib_2}(\omegab, \zb_1) = \prod_{i=1}^d\dfrac{1}{\sqrt{2}\text{exp}(t^i_{\psib_2^2}(\zb_1))},\ s_{\psib_2}(\omegab, \zb_2) = \prod_{i=1}^d\dfrac{1}{\sqrt{2}\text{exp}(t^i_{\psib_2^2}(\zb_2))}
\]
Then we can complete the proof by Theorem \ref{thm:data_dependent_postive_definite}.
\end{proof}

\section{Proof of Theorem \ref{pro:MMD_kernel}}\label{app:MMD_kernel}
\textbf{Theorem \ref{pro:MMD_kernel}} 
\textit{
        By parameterizing the kernel with our KernelNet $\kappa_{\psib}(\zb_1, \zb_2) = \kappa_{\psib_1}(\zb_1, \zb_2) + \kappa_{\psib_2}(\zb_1, \zb_2)$, $\textup{MMD}_{\phib, \psib}(\mathbb{P}, \mathbb{Q})$ is continuous in the weak topology if the following are satisfied:
        \begin{flalign}
           &\mathbb{E}_{\omegab_{\psib_1}}\left[\Vert \omegab_{\psib_1}\Vert^2\right] < \infty, ~    \mathbb{E}_{\omegab_{\psib_2, \zb_1, \zb_2}}\left[\Vert \omegab_{\psib_2, \zb_1, \zb_2}\Vert^2\right] <\infty, ~ \nonumber\\ 
           &\mathbb{E}_{\omegab_{\psib_2, \zb_1, \zb_2}}\left[\Vert \dfrac{\partial \omegab_{\psib_2, \zb_1, \zb_2}}{\partial \zb_1} - \dfrac{\partial \omegab_{\psib_2, \zb_1, \zb_2}}{\partial \zb_2}\Vert_{\mathcal{F}}\right] < \infty,  ~\nonumber\\
           & \sup _{\phib \in \Phib}\Vert h_{\phib} \Vert_{Lip} < \infty     
        \end{flalign}
        where $\Vert \cdot \Vert_{\mathcal{F}}$ denotes the Frobenius norm of a matrix, $h_{\phib}$ is the injective function in MMD-GAN, {\it i.e.}, $\zb=h_{\phib}(\xb)$, and $\Vert h_{\phib} \Vert_{Lip}$ denotes its Lipschitz constant.
}
\begin{proof}
We start from the following Lemma:
\begin{lemma}[\cite{DBLP:conf/nips/ArbelSBG18}]\label{lem:topology}
    Assume the critic functions, which have form:
    \[
        f_{\nub}(\ab) = (\mathbb{E}_{\xb \sim \mathbb{P}} k_{\nub}(\xb, \ab) - \mathbb{E}_{\yb \sim \mathbb{Q}} k_{\nub}(\yb, \ab))/\text{MMD}_{k_{\nub}}(\mathbb{P}, \mathbb{Q})
    \]
    are uniformly bounded and have a common Lipschitz constant: \[\sup _{\ab \in \mathbb{R}^D, \nub \in \Nub} \vert f_{\nub}(\ab)\vert < \infty\, ~ \sup _{\nub \in \Nub} \Vert f_{\nub}(\ab)\Vert_{Lip} < \infty.\] 
    Then $\text{MMD}_{k_{\nub}}(\mathbb{P}, \mathbb{Q})$ is continuous in the weak topology.
    In particular, this holds when $k_{\nub} = \kappa \circ h_{\phi}$ and 
    \[
         \sup _{\zb \in \mathbb{R}^d} \kappa(\zb, \zb) < \infty, ~ \Vert \kappa(\zb_1, \cdot) - \kappa(\zb_2, \cdot)\Vert_{\mathcal{H}_{\kappa}} \leq L_{\kappa} \Vert \zb_1 - \zb_2 \Vert_{\mathbb{R}^d}, ~\sup _{\phib \in \Phib}\Vert h_{\phi} \Vert_{Lip} < \infty
    \]
\end{lemma}
From \eqref{eq:kernel} we know that $\sup_{\zb_1 \in \mathbb{R}^d}\kappa_{\psib}(\zb_1, \zb_1) < \infty$ is naturally satisfied. We prove the second condition in Lemma \ref{lem:topology} here:
\begin{flalign}
    &\Vert \kappa_{\psib}(\zb_1, \cdot) - \kappa_{\psib}(\zb_2, \cdot)\Vert_{\mathcal{H}}\nonumber \\
    \leq&\sqrt{\left<\kappa_{\psib}(\zb_1, \cdot) - \kappa_{\psib}(\zb_2, \cdot), \kappa_{\psib}(\zb_1, \cdot) - \kappa_{\psib}(\zb_2, \cdot)\right>_{\mathcal{H}}}\nonumber \\
    =&\sqrt{\kappa_{\psib}(\zb_1, \zb_1) + \kappa_{\psib}(\zb_2, \zb_2) - 2\kappa_{\psib}(\zb_1, \zb_2)}\nonumber \\
    =&\sqrt{\kappa_{\psib_1}(\zb_1, \zb_1) + \kappa_{\psib_1}(\zb_2, \zb_2)- 2\kappa_{\psib_1}(\zb_1, \zb_2) + \kappa_{\psib_2}(\zb_1, \zb_1) + \kappa_{\psib_2}(\zb_2, \zb_2)- 2\kappa_{\psib_2}(\zb_1, \zb_2)}\nonumber \\
    \leq& \sqrt{2 - 2\kappa_{\psib_1}(\zb_1, \zb_2)} + \sqrt{2 - 2\kappa_{\psib_2}(\zb_1, \zb_2)}\label{eq:pro_5_1}
\end{flalign}
Let start with the second term above, denote $f_2(\tb)=2-\kappa _2(\tb)$, where $\tb = \zb_1-\zb_2$.
\begin{flalign*}
    &\Vert\nabla_{\tb}f_2(\tb)\Vert \\
    =&\Vert \nabla _{\tb}[2-2\kappa_{\psib_2}(\tb)]\Vert\\
    =&\Vert \nabla _{\tb}\{2-2\mathbb{E}_{\omegab_{\psib_2, \tb}}[\cos(\omegab_{\psib_2, \tb}^\intercal\tb)]\}\Vert\\
    =&2\Vert \mathbb{E}_{\omegab_{\psib_2, \tb}}[\sin(\omegab_{\psib_2, \tb}^\intercal\tb)(\omegab_{\psib_2, \tb}^\intercal + \tb \nabla_{\tb}\omegab_{\psib_2, \tb})]\Vert\\
    \leq&2\Vert \mathbb{E}_{\omegab_{\psib_2, \tb}}[\sin(\omegab_{\psib_2, \tb}^\intercal \tb)\omegab_{\psib_2, \tb}]\Vert + 2\Vert \mathbb{E}_{\omegab_{\psib_2, \tb}}[\sin(\omegab^\intercal \tb)\tb \nabla_{\tb}\omegab_{\psib_2, \tb}]\Vert\\
    \leq&2\mathbb{E}_{\omegab_{\psib_2, \tb}}(\Vert \tb\Vert \Vert \omegab_{\psib_2, \tb} \Vert^2) + 2\mathbb{E}_{\omegab_{\psib_2, \tb}}(\Vert\tb\Vert \Vert\nabla_{\tb} \omegab_{\psib_2, \tb}\Vert_{\mathcal{F}})\\
    =&2\mathbb{E}_{\omegab_{\psib_2, \tb}}[\Vert \tb \Vert (\Vert {\omegab_{\psib_2, \tb}}\Vert^2 + \Vert \nabla_{\tb} \omegab_{\psib_2, \tb} \Vert_{\mathcal{F}})]
\end{flalign*}
If $2\mathbb{E}_{\omegab_{\psib_2, \tb}}\left[\Vert {\omegab_{\psib_2, \tb}}\Vert^2\right]\leq c_1$ and $2\mathbb{E}_{\omegab_{\psib_2, \tb}}\left[\Vert \nabla_{\tb} \omegab_{\psib_2, \tb} \Vert\right]\leq c_2$, for constants $c_1$, $c_2$, then we have $\Vert\nabla_{\tb}f_2(\tb)\Vert\leq c\Vert \tb \Vert$, where $c=c_1+c_2$.
\begin{flalign*}
   \Vert f_2(\tb) \Vert  &= \Vert \int \nabla_{\tb} f_2(\tb)d\tb\Vert = \int \Vert \nabla_{\tb} f_2(\tb)\Vert d\tb\\
    &\leq c\int \Vert \tb\Vert d\tb =c\int \sqrt{\sum_{i=1}^d\tb_i^2}d\tb\\
    &\leq c\sum_{i=1}^d\int \vert \tb_i\vert d\tb_i\\
    &\leq c\sum_{i=1}^d\dfrac{\tb_i^2}{2} =\dfrac{c}{2}\Vert \tb \Vert^2
\end{flalign*}
Thus we can conclude, if \[\mathbb{E}_{\omegab_{\psib_2, \tb}}\left[\Vert \omegab_{\psib_2, \tb}\Vert^2\right] <\infty, ~  \mathbb{E}_{\omegab_{\psib_2, \tb}}\left[\Vert \nabla_{\tb} \omegab_{\psib_2, \tb} \Vert_{\mathcal{F}}\right] < \infty\] then $\sqrt{f_2(\tb)}\leq \sqrt{\dfrac{c}{2}}\Vert \tb \Vert$ holds. Similar result can be easily get for the first term in \eqref{eq:pro_5_1}. Then we can conclude if:
\begin{equation}\label{eq:pro_5_2}
    \mathbb{E}_{\omegab_{\psib_1}}\left[\Vert \omegab_{\psib_1}\Vert^2\right] < \infty, ~    \mathbb{E}_{\omegab_{\psib_2, \tb}}\left[\Vert \omegab_{\psib_2, \tb}\Vert^2\right] <\infty, ~  \mathbb{E}_{\omegab_{\psib_2, \tb}}\left[\Vert \nabla_{\tb} \omegab_{\psib_2, \tb} \Vert_{\mathcal{F}}\right] < \infty
\end{equation}
then
\begin{equation*}
    \Vert \kappa_{\psib}(\zb_1, \cdot) - \kappa_{\psib}(\zb_2, \cdot)\Vert_{\mathcal{H}}\leq L_{\kappa}\Vert \tb\Vert = L_{\kappa}\Vert \zb_1 -\zb_2 \Vert
\end{equation*}
For some constant $L_{\kappa}$. Because $\tb = \zb_1 - \zb_2$, we have:
\begin{equation}\label{eq:pro_5_3}
    \nabla_{\tb}\omegab_{\psib_2, \tb} = \dfrac{\partial \omegab_{\psib_2, \zb_1, \zb_2}}{\partial \zb_1}\dfrac{\partial \zb_1}{\partial \tb} + \dfrac{\partial \omegab_{\psib_2, \zb_1, \zb_2}}{\partial \zb_2}\dfrac{\partial \zb_2}{\partial \tb} = \dfrac{\partial \omegab_{\psib_2, \zb_1, \zb_2}}{\partial \zb_1} - \dfrac{\partial \omegab_{\psib_2, \zb_1, \zb_2}}{\partial \zb_2}
\end{equation}
Substitute \eqref{eq:pro_5_3} into \eqref{eq:pro_5_2}, Theorem \ref{pro:MMD_kernel} is proved.
\end{proof}

\section{Details on Proposition \ref{pro:SMMD_kernel}}\label{app:SMMD_kernel}
\textbf{Proposition \ref{pro:SMMD_kernel}} 
\textit{
    With the proposed data-dependent KernelNet \eqref{eq:kernelp}, the SMMD-DK framework  can be formulated and simplified as:
    \[
    \textup{SMMD-DK}_{\phib, \psib,\lambda}(\mathbb{P}, \mathbb{Q}):= \sigma_{\phib, \psib,\lambda}\textup{MMD}_{\phib, \psib}(\mathbb{P}, \mathbb{Q}),
    \]
    where
    {\small
    \begin{align*}
        \sigma_{\phib, \psib, \lambda}:=\bigg\{&\lambda + 1 + \mathbb{E}_{\xb \sim \mathbb{P}}\{\mathbb{E}_{\omegab_{\psib_2, \zb, \zb}}[\Vert\omegab_{\psib_2, \zb, \zb}\Vert^2]\Vert\nabla h_{\phib}(\xb)\Vert _{\mathcal{F}}\} +\mathbb{E}_{\xb \sim \mathbb{P}}\{\mathbb{E}_{\omegab_{\psib_1}}[\Vert\omegab_{\psib_1}\Vert^2]\Vert\nabla h_{\phib}(\xb)\Vert _{\mathcal{F}}\}\bigg\}^{-1/2},
    \end{align*}
        } 
    and $\zb = h_{\phib}(\xb)$.
}

In our data-dependent kernel setting, $k=\kappa_{\psib}\circ h_{\phib}$, hence we write $k(\yb, \zb)=\kappa_{\psib}(h_{\phib}(\yb), h_{\phib}(\zb))$, where $h_{\phib}$ is the discriminator, and $\kappa_{\psib}$ is the proposed data-dependent kernel.
For our data-dependent kernel, we know that
\[\int \kappa_{\psib, \phib}(\xb, \xb) d\mathbb{P}(\xb) =1\] 
from \eqref{eq:kernel}.
\begin{flalign*}
    ~&\sum_{i=1}^d\int \dfrac{\partial^2\kappa_{\psib}(\yb_{1}, \yb_{2})}{\partial\yb_{1i} \partial\yb_{2i}}\vert_{(\yb_{1}, \yb_{2})=(\xb,\xb)} d\mathbb{P}(\xb)\\
    =& \sum_{i=1}^d\int \dfrac{\partial^2\kappa_{\psib_1}(\yb_1, \yb_2)}{\partial\yb_{1i} \partial\yb_{2i}}\vert_{(\yb_1, \yb_2)=(\xb,\xb)} d\mathbb{P}(\xb) +  \sum_{i=1}^d\int \dfrac{\partial^2\kappa_{\psib_2}(\yb_{1},\yb_{2})}{\partial\yb_{1i} \partial\yb_{2i}}\vert_{(\yb_{1},\yb_{2})=(\xb,\xb)} d\mathbb{P}(\xb)
\end{flalign*}
where $\zb_1 = h_{\phib}(\yb_1)$ and $\zb_2 = h_{\phib}(\yb_2)$. For clearness, we will write $\omegab_{\psib_2}$ instead of $\omegab_{\psib_2, \zb_1, \zb_2}$, but please keep in mind that $\omegab_{\psib_2}$ is dependent on the input data. Because of the reparameterization trick, we can write $\mathbb{E}_{\epsilon \sim \mathcal{N}(\mathbf{0}, \mathbf{I})}(\cdot)$ instead of $\mathbb{E}_{\omegab_{\psib, \zb_1, \zb_2}}(\cdot)$,
\begin{flalign*}
    ~& \sum_{i=1}^d\int \dfrac{\partial^2\kappa_{\psib_2}(\yb_{1},\yb_{2})}{\partial\yb_{1i} \partial\yb_{2i}}\vert_{(\yb_{1},\yb_{2})=(\xb,\xb)} d\mathbb{P}(\xb)\\
    =&\sum_{i=1}^d\int \dfrac{\partial^2\mathbb{E}_{\epsilon \sim \mathcal{N}(\mathbf{0}, \mathbf{I})}\{\cos \{\omegab_{\psib_2}^\intercal[h_{\phib}(\yb_1) - h_{\phib}(\yb_2)]\}\}}{\partial\yb_{1i}\partial\yb_{2i}}\vert_{(\yb_1, \yb_2)=(\xb,\xb)} d\mathbb{P}(\xb)\\
    =& \sum_{i=1}^d\int \int \dfrac{\partial}{\partial \yb_{2i}}\sin \{\omegab_{\psib_2}^\intercal[h_{\phib}(\yb_1) - h_{\phib}(\yb_2)]\} \\
    &~~~~~~~~~~~~~~~~~~~~~~~~~~~~~~~~~\{\dfrac{\partial\omegab_{\psib_2}} {\partial \yb_{1i}}^\intercal[h_{\phib}(\yb_1) - h_{\phib}(\yb_2)] + \omegab_{\psib_2}^\intercal\dfrac{\partial h_{\phib}(\yb_1)}{\partial \yb_{1i}}\} d\mub(\epsilonb) \vert_{(\yb_1, \yb_2)=(\xb,\xb)} d\mathbb{P}(\xb) \\
    =&\sum_{i=1}^d\int \int  -\cos \{\omegab_{\psib_2}^\intercal[h_{\phib}(\yb_1)-h_{\phib}(\yb_2)]\}\{\dfrac{\partial \omegab_{\psib_2}}{\partial \yb_{2i}}^\intercal[h_{\phib}(\yb_1) - h_{\phib}(\yb_2)] + \omegab_{\psib_2}^\intercal[-\frac{\partial h_{\phib}(\yb_2)}{\partial \yb_{2i}}]\}\\ &\times \{\dfrac{\partial \omegab_{\psib_2}}{\partial \yb_{1i}}^\intercal[h_{\phib}(\yb_1)-h_{\phib}(\yb_2)] + \omegab_{\psib_2}^\intercal\dfrac{\partial h_{\phib}(\yb_1)}{\partial \yb_{1i}}\} \\&+ \sin \{\omegab_{\psib_2}^\intercal[h_{\phib}(\yb_1)-h_{\phib}(\yb_2)]\}\dfrac{\partial}{\partial \yb_{2i}}\{\dfrac{\omegab_{\psib_2}}{\partial \yb_{1i}}^\intercal[h_{\phib}(\yb_1) - h_{\phib}(\yb_2)] + \omegab_{\psib_2}^\intercal \dfrac{\partial h_{\phib}(\yb_1)}{\partial \yb _{1i}}\}
    d\mub(\epsilonb)\vert_{(\yb_1, \yb_2)=(\xb,\xb)}  d\mathbb{P}(\xb)\\
    =&\sum_{i=1}^d\int \int [\omegab_{\psib_2, \zb, \zb}^\intercal \dfrac{\partial h_{\phib}(\xb)}{\partial \xb_i}]^2  d\mub(\epsilonb)  d\mathbb{P}(\xb)\\
\end{flalign*}
where $\zb = h_{\phib}(\xb)$.
In our experiment, the output dimension of discriminator $h_{\phi}$ is set to be 1. Then the result becomes:
\begin{equation*}
    \sum_{i=1}^d\int  \dfrac{\partial^2\kappa_{\omegab_{\psib_2, \zb_1,\zb_2}}(\yb_1, \yb_2)}{\partial\yb_{1i} \partial\yb_{2i}}\vert_{(\yb_1, \yb_2)=(\xb,\xb)}  d\mathbb{P}(\xb)\\
    =\mathbb{E}_{\xb \sim \mathbb{P}}\{\mathbb{E}_{\omegab_{\psib_2, \zb, \zb}}[\Vert\omegab_{\psib_2, \zb, \zb}\Vert^2] \Vert \nabla h_{\phib}(\xb)\Vert _{\mathcal{F}}\}
\end{equation*}
Similarly, we have:
\begin{equation*}
    \sum_{i=1}^d\int\dfrac{\partial^2\kappa_{\omegab_{\psib_1}}(\yb_1, \yb_2)}{\partial\yb_{1i} \partial\yb_{2i}}\vert_{(\yb_1, \yb_2)=(\xb,\xb)}d\mathbb{P}(\xb)\\
    =\mathbb{E}_{\xb \sim \mathbb{P}}\{\mathbb{E}_{\omegab_{\psib_1}}[\Vert\omegab_{\psib_1}\Vert^2] \Vert \nabla h_{\phib}(\xb)\Vert _{\mathcal{F}}\}
\end{equation*}
Hence:
\begin{align}
    &\sum_{i=1}^d\int \dfrac{\partial^2\kappa_{\omegab}(\yb_1, \yb_2)}{\partial\yb_{1i} \partial\yb_{2i}}\vert_{(\yb_1, \yb_2) =(\xb,\xb)}d\mathbb{P}(\xb) \nonumber\\
    &= \mathbb{E}_{\xb \sim \mathbb{P}}\{\{\mathbb{E}_{\omegab_{\psib_2, \zb, \zb}}[\Vert\omegab_{\psib_2, \zb, \zb}\Vert^2] + \mathbb{E}_{\omegab_{\psib_1}}[\Vert\omegab_{\psib_1}\Vert^2] \}\Vert\nabla h_{\phib}(\xb)\Vert _{\mathcal{F}}\}
\end{align}

\section{Experimental Settings}\label{app:setting}
For a fair comparison, all the models are evaluated under the same architecture on each dataset. Our model architectures follow \cite{DBLP:conf/nips/ArbelSBG18}. For CIFAR-10 and STL-10, we use an architecture with a 7-layer convolutional neural network (CNN) as the discriminator and a 4-layer CNN as the generator. For CelebA, we use a 5-layer CNN discriminator and a 10-layer ResNet generator. For ImageNet, our generator and discriminator are both 10-layer ResNets. 
The output dimension of discriminator is set to be 1 for all the models, except that it is set to 16 when repulsive loss is used. Inputs of the generator are sampled from a uniform distribution $\mathcal{U}[-1,1]^{128}$. We use two 3-layer fully-connected neural networks to parameterize $\omegab_{\psib_1}$ and $\omegab_{\psib_2, \zb_1, \zb_2}$. For each neural network, there are 16 neurons in every hidden layer when the discriminator's output dimension is 1, and 64 neurons when the discriminator's output dimension is 16. 

Spectral normalization \cite{DBLP:conf/iclr/MiyatoKKY18} is used in most of the models except WGAN-GP and SMMD-GAN, spectral parameterization \cite{DBLP:conf/nips/ArbelSBG18} is used in SN-SMMD-GAN-DK. Note that in Rep-GAN-DK, we scale the weight after spectral normalization by a constant chosen from \{0.5, 1, 2\} based on hyper-parameter tuning, which is similar to \cite{wang2018improving}.

Adam optimizer \cite{kingma2014adam} with batch size of 64 is used in all the experiments. Learning rates of generator and discriminator are selected from $\{0.0001, 0.0002\}$. At every update step, 1024 samples of $\omegab_{\psib_1}$ and $\omegab_{\psib_2}$ are used to compute the values of the kernel function. We set $\alpha_1=0,\ \alpha_2=0.1$ in \eqref{eq:mmd_dk}, $\alpha_1=0,\ \alpha_2=0.1$ in \eqref{eq:smmd_dk} and $\alpha_1=0.1,\ \alpha_2=0.1$ in \eqref{eq:mmd_rep_dk} respectively. 

In SN-MMD-GAN-DK and SN-SMMD-GAN-DK, we update discriminator and KernelNet 5 steps for every generator update. $\zeta$ in \eqref{eq:real_smmd_obj} is selected from $\{1,2,5\}$. Ratio of learning rate of KernelNet to learning rate of generator is selected from $\{0.01, 0.005\}$. The hyper-parameters of Adam optimizer are set to be $\beta_1=0.5$, $\beta_2=0.9$. Models are trained for 150,000 generator update steps for CIFAR-10, STL-10 and CelebA, 200,000 generator update steps for ImageNet.

In Rep-GAN-DK, we update the discriminator and KernelNet one step for every generator update, the learning rate of KernelNet is set to be half of the generator. The hyper-parameters of Adam optimizer are set to be $\beta_1=0.5$, $\beta_2=0.999$. $\eta$ in \eqref{eq:rep_loss} is selected from $\{0, 0.5, 1\}$. Models are trained for 200,000 generator update steps for CIFAR-10, STL-10 and CelebA, 300,000 generator update steps for ImageNet. 

We report the standard Fréchet Inception Distance (FID) \cite{DBLP:journals/corr/HeuselRUNKH17} and Inception Score (IS) \cite{SalimansGZCRC:NIPS16}. They are computed using 100,000 samples on CIFAR-10, Stl-10 and ImageNet datasets, while 50,000 samples are used on CelebA due to the GPU memory limitation. During the training process, we decrease the learning rate based on the relative KID test \cite{DBLP:journals/corr/BounliphoneBBAG15}. The frequency of decreasing the learning rate are based on hyper-parameter tuning.

\clearpage
\section{Extra Experiments Results}\label{app:results}
\subsection{MMD-GAN with KernelNet}
Some generated images of our proposed methods are shown in the figures below. 
\begin{figure}[h!]
    \vskip 0.1in
    \centering
    \subfigure[CIFAR-10($32 \times 32$)]{\includegraphics[width=0.23\linewidth]{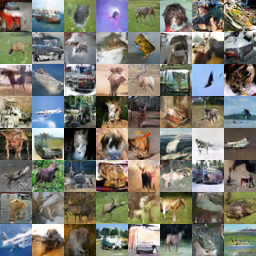}
    \label{fig:cifar_32}}
    \subfigure[STL-10($48 \times 48$)]{\includegraphics[width=0.23\linewidth]{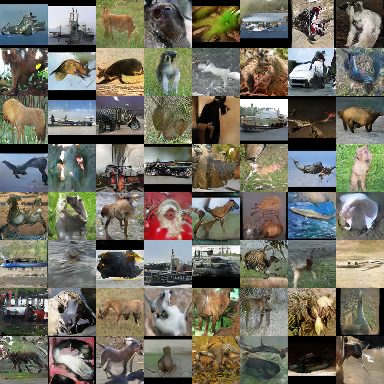}
    \label{fig:stl_32}}
    \subfigure[ImageNet ($64 \times 64$)]{\includegraphics[width=0.23\linewidth]{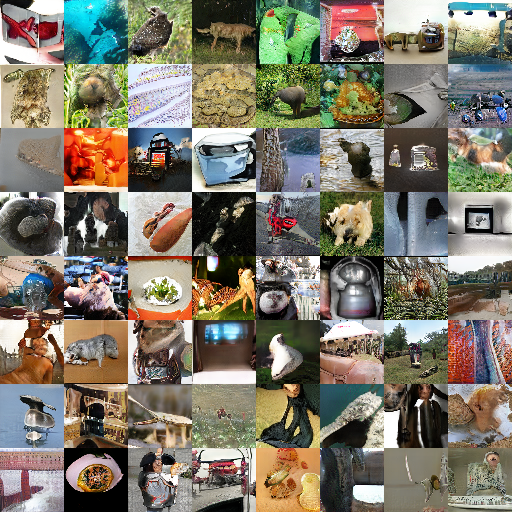}
    \label{fig:imagenet_64}}
    \subfigure[CelebA ($160 \times 160$)]{\includegraphics[width=0.23\linewidth]{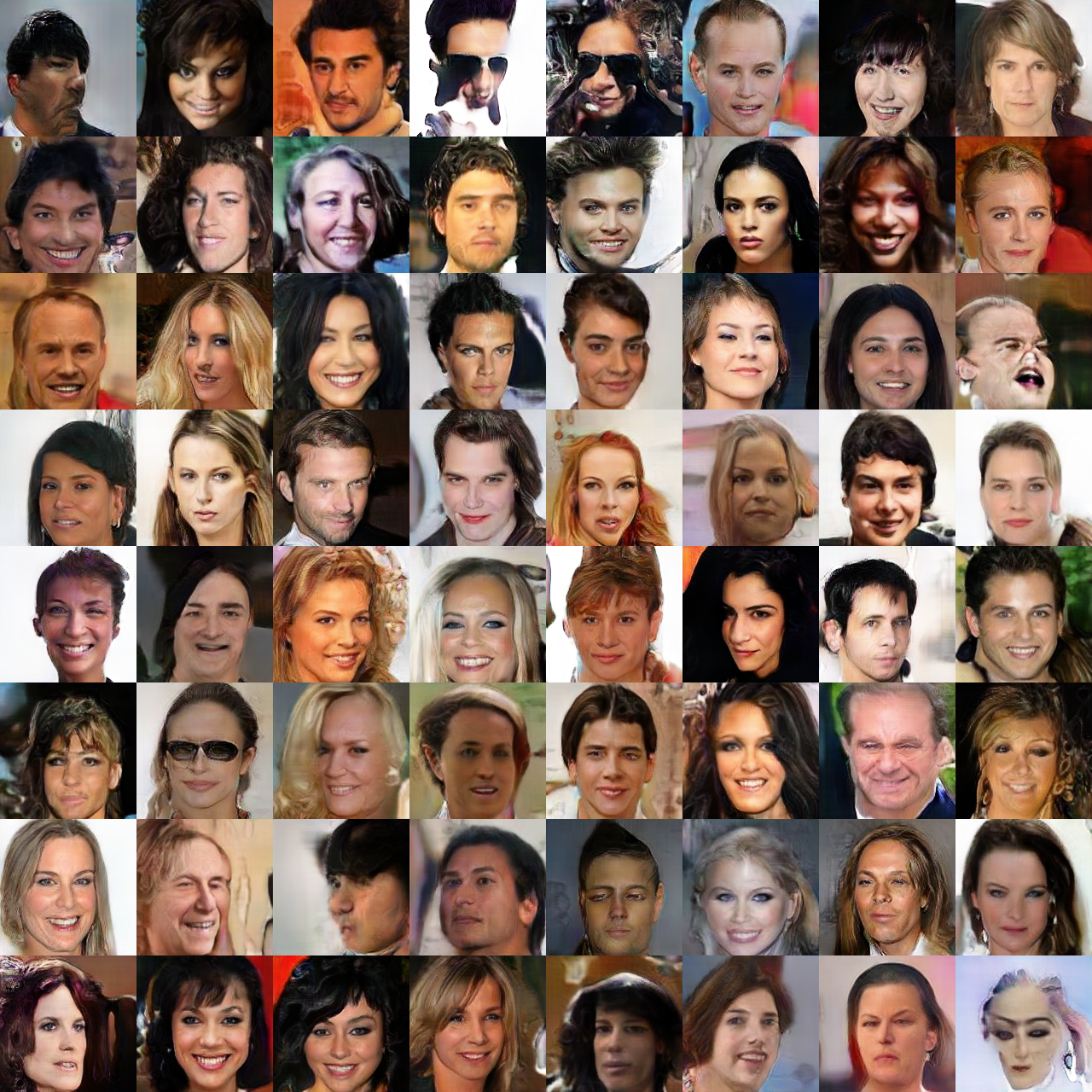}
    \label{fig:celeba_160}}
    \caption{Generated images of SN-SMMD-GAN-DK.}
    \vskip -0.1in
\end{figure}
\begin{figure}[h!]
    \vskip 0.1in
    \centering
    \subfigure[CIFAR-10($32 \times 32$)]{\includegraphics[width=0.23\linewidth]{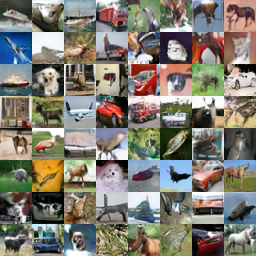}
    \label{fig:cifar_32}}
    \subfigure[STL-10($48 \times 48$)]{\includegraphics[width=0.23\linewidth]{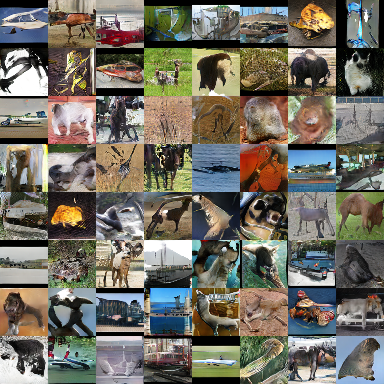}
    \label{fig:stl_32}}
    \subfigure[ImageNet ($64 \times 64$)]{\includegraphics[width=0.23\linewidth]{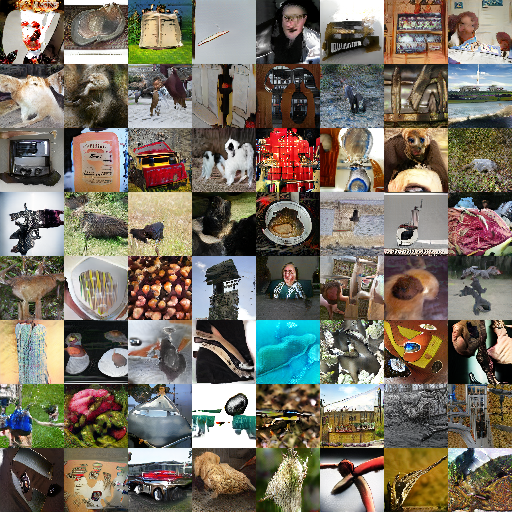}
    \label{fig:imagenet_64}}
    \subfigure[CelebA ($160 \times 160$)]{\includegraphics[width=0.23\linewidth]{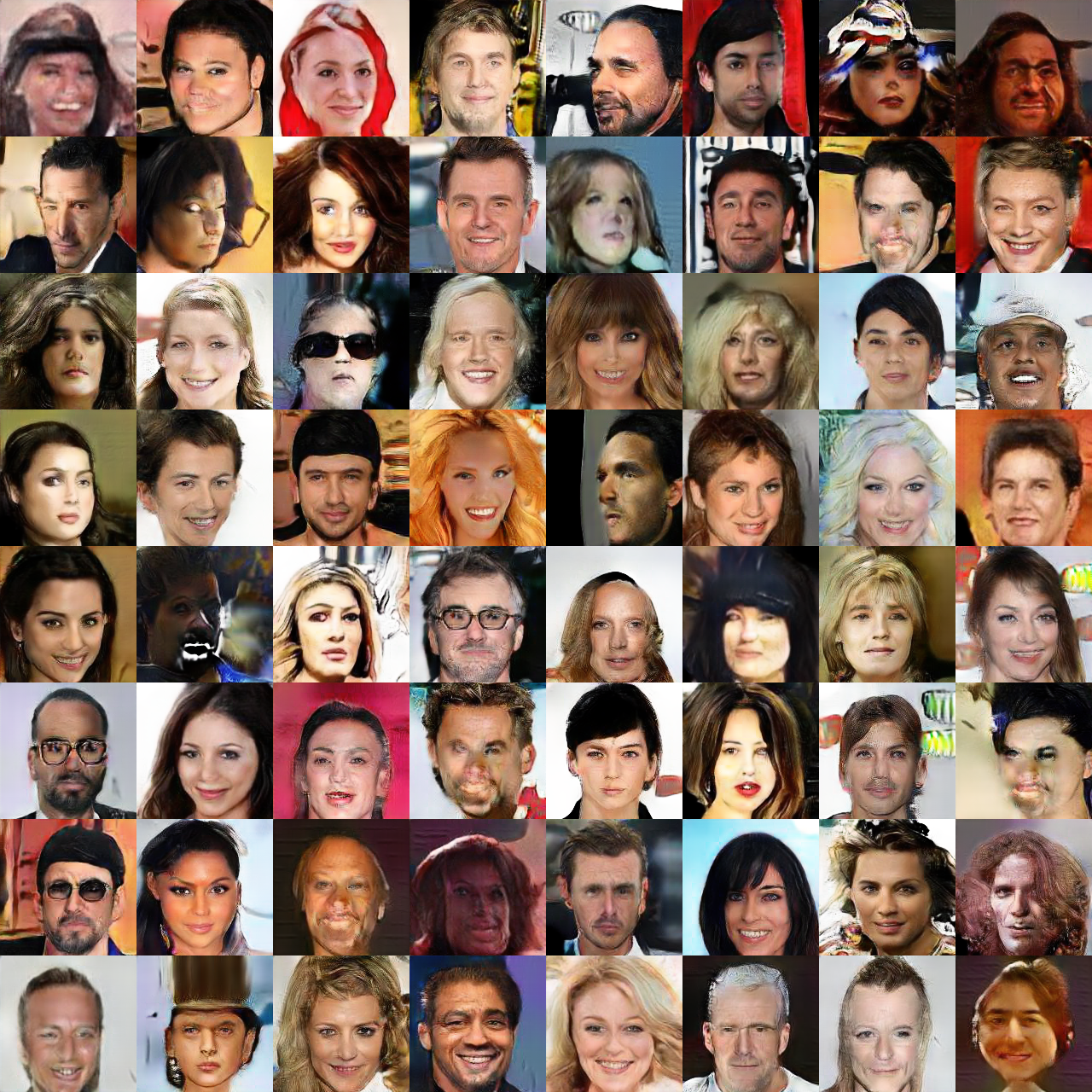}
    \label{fig:celeba_160}}
    \caption{Generated images of Rep-GAN-DK.}\label{fig:generated_img_rep_dk}
    \vskip -0.1in
\end{figure}

There are many different ways to construct data-dependent distribution. In the main text, we provide a simple way whose resulting kernel is guaranteed to be positive definite. For example, We can set $t_{\psib_2^1}$ and $t_{\psib_2^2}$'s input to be $\left[ \zb_1, \zb_2\right]$, i.e. concatenation of $\zb_1, \zb_2$, which leads to
\begin{align}\label{eq:another_way_to_construct_kernel}
        &\omegab_{\psib_2, \zb_1, \zb_2} = \mub_{\psib_2,\zb_1,\zb_2} + \epsilonb \odot \sigmab_{\psib_2, \zb_1, \zb_2}, \text{ where } \epsilonb \sim  \mathcal{N}(0, 1)\\
        &\mub_{\psib_2, \zb_1, \zb_2}=t_{\psib_2^1}(\left[ \zb_1, \zb_2\right]) + t_{\psib_2^1}(\left[ \zb_2, \zb_1\right]),  \sigmab_{\psib_2, \zb_1, \zb_2}=\text{exp}(t_{\psib_2^2}(\left[ \zb_1, \zb_2\right]) +  t_{\psib_2^2}(\left[ \zb_2, \zb_1\right])),  \nonumber
\end{align}
Although the positive definiteness may not be guaranteed in this case, the experimental results are still competitive. We use DK-\ROM{2} to denote the data-dependent kernel following \eqref{eq:another_way_to_construct_kernel}, and provide the comparison with previous construction \eqref{eq:kernel_part_2} here.

\begin{table*}[hb!]
    \vspace{-0.2cm}
    \caption{Comparison of Different Kernel Construction.}
    \vskip 0.1in
    \begin{center}
        \begin{sc}
        \begin{adjustbox}{scale=0.7}
            \begin{tabular}{lcccccccc}
            \toprule
            &\multicolumn{2}{c}{CIFAR-10} & \multicolumn{2}{c}{STL-10} &\multicolumn{2}{c}{CelebA} & \multicolumn{2}{c}{ImageNet}\\
            & FID $(\downarrow)$ & IS $(\uparrow)$ & FID $(\downarrow)$ & IS $(\uparrow)$ & FID $(\downarrow)$ & IS $(\uparrow)$ & FID $(\downarrow)$ & IS $(\uparrow)$\\
            \midrule
            Rep-GAN & $16.7$ & $8.0$ & $36.7$ & $\mathbf{9.4}$ & $16.8 \pm 0.1 $ & $2.9 \pm 0.1$ & $31.0 \pm 0.1$ & $11.5 \pm 0.1$\\
            Rep-GAN-DK (ours) &$\mathbf{14.6 \pm 0.1}$ & $\mathbf{8.2 \pm 0.1}$ &$\mathbf{33.4\pm 0.1}$&$9.3 \pm 0.1$& $16.5 \pm 0.1$ & $2.9 \pm 0.1$ & $\mathbf{30.1 \pm 0.1}$ & $\mathbf{11.7 \pm 0.1}$ \\
            Rep-GAN-DK-\ROM{2} (ours) & $15.2 \pm 0.1 $ & $8.1 \pm 0.1$ & $34.9 \pm 0.1$ &$9.3 \pm 0.1$& $\mathbf{16.1 \pm 0.1}$ & $\mathbf{2.9 \pm 0.1}$ & $30.5 \pm 0.1$ & $11.7 \pm 0.1$\\
            \bottomrule
            \end{tabular}  
        \end{adjustbox}
        \end{sc}
    \end{center}
    \vskip -0.1in
\end{table*}

\clearpage
\subsection{Extra Experiments on Info-VAE}
\begin{figure}[t!]
    \vskip 0.1in
	\centering
	\includegraphics[width=.49\linewidth]{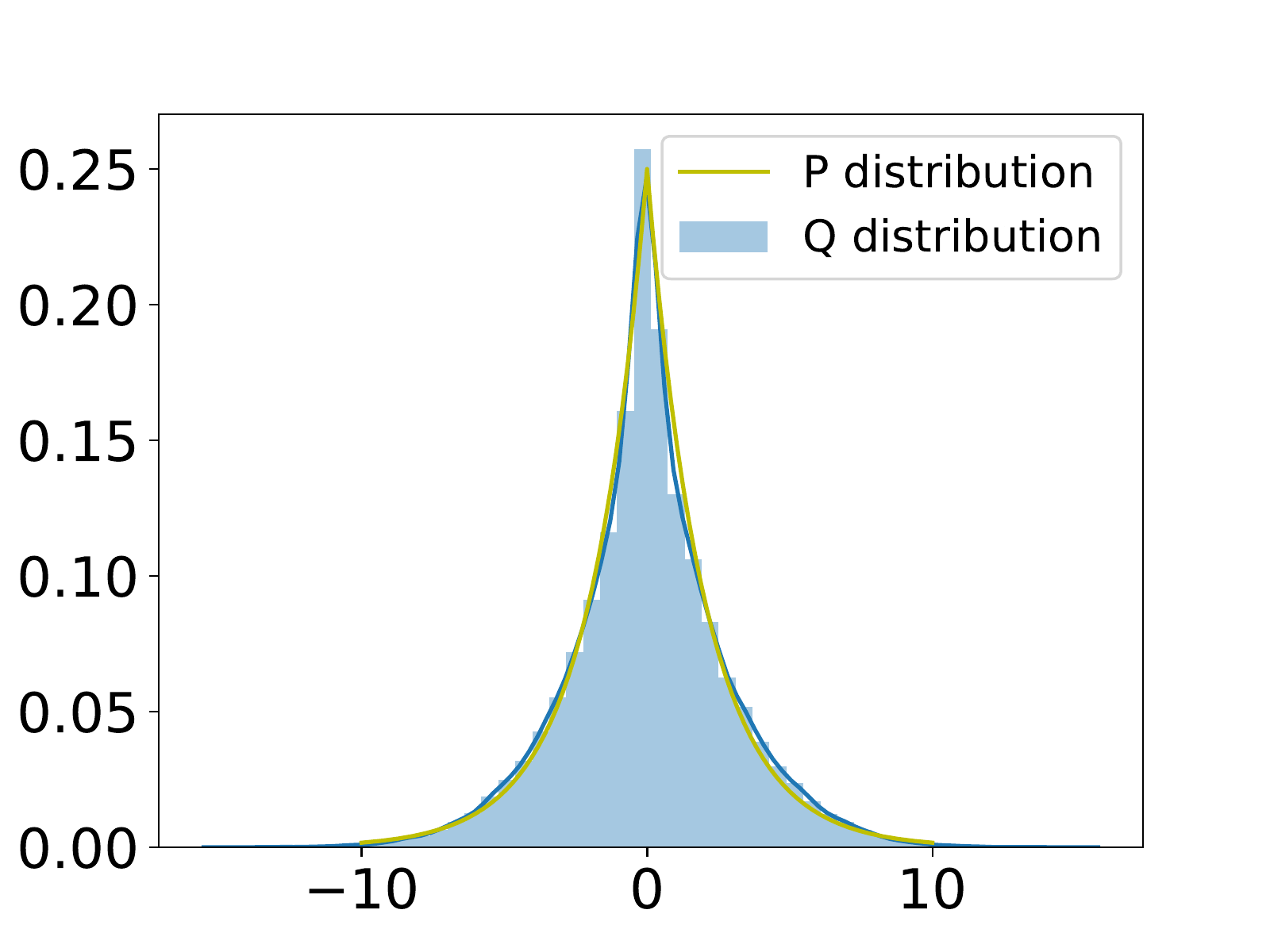}
	\includegraphics[width=.49\linewidth]{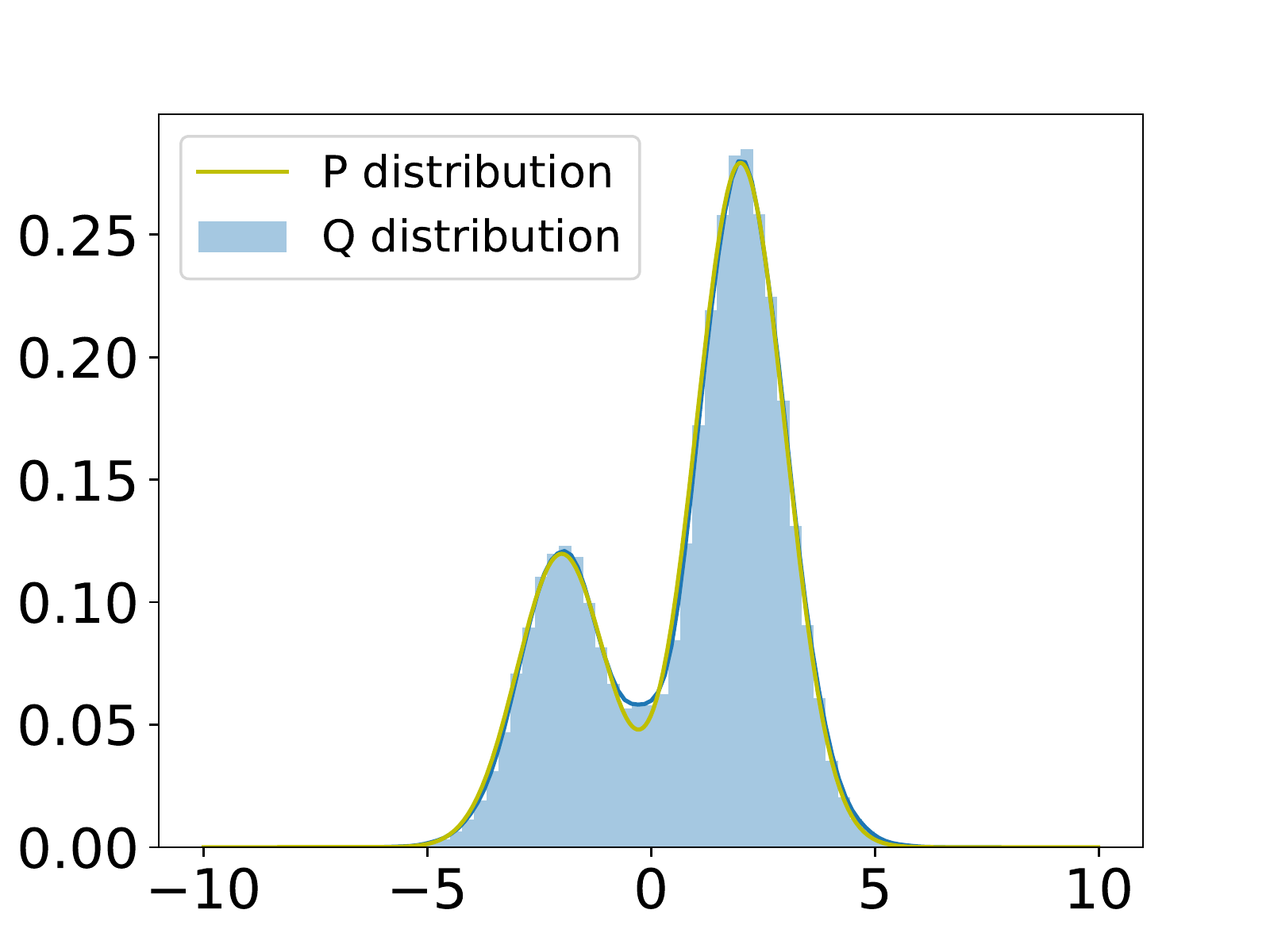}
    \caption{Learning to sample from two target distributions: $\text{Laplace}(0,2)$ (left) and Gaussian mixture $0.3\mathcal{N}(-2, 1) + 0.7\mathcal{N}(2, 1)$ (right). $P$-distribution denotes the ground truth; $Q$-distribution denotes the approximated density by samples.}
    \label{fig:syn_samples}
    \vskip -0.1in
\end{figure}

\paragraph{Multi-modal distribution sampling}
We first illustrate the implicit encoder can learn latent variable with multi-mode distributions. This is done by removing the decoder and only training the encoder, which essentially learns a parametric sampler. We use a 3-layer fully-connected neural network with 20 hidden units as the encoder, whose inputs are Gaussian noises. Figure \ref{fig:syn_samples} plots the learned distributions estimated by samples on two target distribution, which can perfectly generates multi-mode samples.

\paragraph{Implicit VAE}

Next, we test our Implicit Info-VAE model on the MNIST dataset \cite{DBLP:conf/icml/SalakhutdinovM08} to learn an implicit VAE model. We use a fully-connected neural network with 1 hidden layer for both encoder and decoder, whose hidden units are set to 400. $\omegab_{\psib_1}$ and $\omegab_{\psib_2, \zb_1, \zb_2}$ are parameterized by DNNs consisting of 2 fully connected hidden layers with 32 hidden units. Bernoulli noises are injected into the encoder by using dropout with a dropout rate of 0.3. The latent dimension is 32. 
The models are trained for 300 epochs. Stochastic gradient descent (SGD) with momentum of 0.9 is used with a batch size of 32. We sample 32 $\zb$ for every $\xb$. The learning rate for the encoder and decoder is 0.002, while it is 0.001 for kernel learning. At every step, we sample 512 random features from the spectral distribution. 

For fair evaluation, we follow \cite{DBLP:journals/corr/WuBSG16} and use Annealed Importance Sampling (AIS) to approximate the negative log-likelihood (NLL). 10 independent AIS chains are used, each of which have 1000 intermediate distributions. The final results are computed using 5000 random sampled test data. The results are shown in Table \ref{tab:mnist_nll_2}, where we compare with related models including: VAE (vanilla VAE from \cite{KingmaW:ICLR14}), Stein-VAE (amortized SVGD from \cite{FengWL:UAI17}), SIVI (Semi-Implicit VAE from \cite{DBLP:conf/icml/YinZ18}), Spectral (implicit VAE with spectral method for gradient estimation from \cite{shi2018spectral}) and Info-VAE \cite{DBLP:journals/corr/ZhaoSE17b}. 

We denote our Implicit Info-VAE with Stein gradient estimator with objective \eqref{eq:infovae_obj} as Info-IVAE. The models with objective \eqref{eq:infovae_obj_mmd} are denoted as Info-IVAE-RBF, Info-IVAE-IKL and Info-IVAE-DK, where the MMD regularizers are computed by RBF kernel, implicit kernel without data-depedent component and data-dependent KernelNet respectively. 

\begin{figure}[t!]
    \centering
    \vskip 0.1in
    \subfigure[Reconstruction]{
        \includegraphics[width=0.4\linewidth]{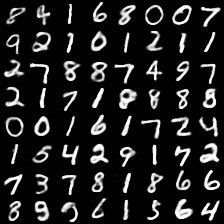}
        \label{fig:rec_mnist}}
    \subfigure[Generation]{
        \includegraphics[width=0.4\linewidth]{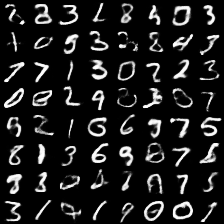}
        \label{fig:gen_mnist}}
    \caption{Reconstructed and generated images on MNIST. }
    \label{fig:mnist_images}
    \vskip -0.1in
\end{figure}

Note that some models have also reported scores related to NLL in their original paper under different settings, which are not directly comparable to ours. For fair comparisons, we use the same encoder-decoder structure and rerun all the models. Our model obtains the best NLL score among all the models. Some reconstructed images and generated images of our model are shown in Firgure \ref{fig:mnist_images}.

\begin{table*}[ht!]
    \vspace{-0.2cm}
    \caption{Negative log-likelihood on the binarized MNIST dataset.}
    \label{tab:mnist_nll_2}  
    \vskip 0.1in
    \begin{center}
    \begin{sc}
        \begin{adjustbox}{scale=0.7,center}
            \begin{tabular}{lccccccccc}
            \toprule
             Model&VAE&Stein-VAE&Spectral&SIVI&Info-VAE&Info-IVAE&Info-IVAE-RBF&Info-IVAE-IKL&Info-IVAE-DK \\
             \midrule
             NLL~$\downarrow$ &90.32&88.85&89.67&89.03&88.89&89.79&88.24&88.21&\textbf{88.16}\\
             \bottomrule
            \end{tabular}
        \end{adjustbox}
    \end{sc}
    \end{center}
    \vskip -0.1in
\end{table*}

\begin{figure}[tb!]
    \vskip 0.1in
    \centering
    \includegraphics[width=0.4\linewidth]{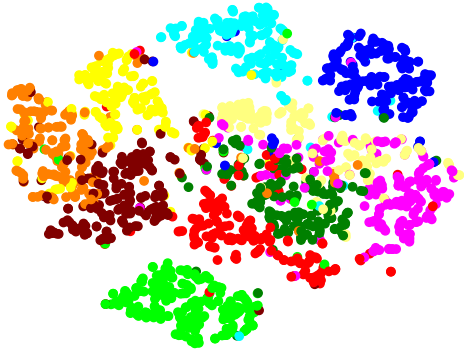}
    \includegraphics[width=0.4\linewidth]{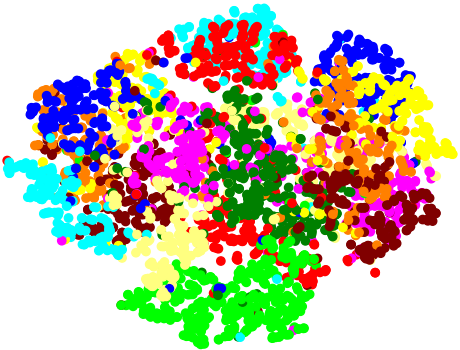}
    \caption{T-SNE visualization of learned latent variables on MNIST. Left and right figures correspond to the implicit kernel with/without data-dependent component, respectively.}
    \label{fig:latent_tsne}
    \vskip -0.1in
\end{figure}
We also plot the $t$-SNE visualization of latent variables learned by Info-IVAE-IK and Info-IVAE-DK in Figure \ref{fig:latent_tsne}. From the figure we can see that latent variables learned using data-dependent kernel looks more separable than implicit kernel without the data-dependent part. 

\begin{figure}[htb!]
    \centering
    \includegraphics[width=0.4\textwidth]{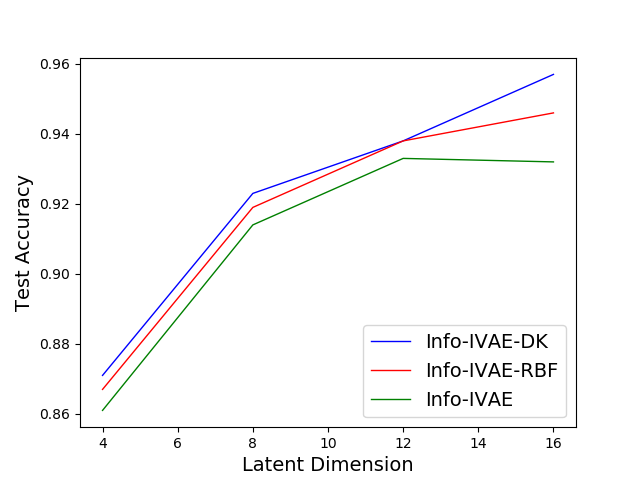}
    \caption{Semi-supervised experiment on Mnist dataset}
    \label{fig:semi}
\end{figure}

We evaluate the latent variables learned by Info-IVAE with different kernels following \cite{kingma2014semi}. After we finish training Info-IVAE models, we generate latent features using the encoders. Then we train a SVM on these latent features. More informative latent variables should lead to better classification performance, the results are shown in Figure \ref{fig:semi}.
\end{document}